\documentclass[letterpaper]{article} % DO NOT CHANGE THIS
\usepackage{aaai2026}  % DO NOT CHANGE THIS
\usepackage{times}  % DO NOT CHANGE THIS
\usepackage{helvet}  % DO NOT CHANGE THIS
\usepackage{courier}  % DO NOT CHANGE THIS
\usepackage[hyphens]{url}  % DO NOT CHANGE THIS
\usepackage{graphicx} % DO NOT CHANGE THIS
\usepackage{natbib}  % DO NOT CHANGE THIS AND DO NOT ADD ANY OPTIONS TO IT
\usepackage{caption} % DO NOT CHANGE THIS AND DO NOT ADD ANY OPTIONS TO IT
\usepackage{multirow}
\usepackage{algorithm}
\usepackage{algorithmic}
\usepackage{appendix}
\usepackage{makecell}
\usepackage{amsfonts}

\newcommand{\ourmethod}{LLM2Comp}
\newcommand{\ourmethodrc}{LLM2Comp$_{RC}$}
\newcommand{\ourmethodnll}{LLM2Comp$_{NLL}$}
\newcommand{\ourmethodkl}{LLM2Comp$_{KL}$}
\newcommand{\llmtovec}{LLM2Vec}
\newcommand{\lamatovec}{Llama2Vec}

\usepackage{natbib}  % DO NOT CHANGE THIS AND DO NOT ADD ANY OPTIONS TO IT
\usepackage{caption} % DO NOT CHANGE THIS AND DO NOT ADD ANY OPTIONS TO IT
\usepackage{enumitem}
\usepackage{times}
\usepackage{latexsym}
\usepackage{booktabs}
\usepackage{tcolorbox}
\usepackage{microtype}
\usepackage{times}
\usepackage{latexsym}
\usepackage{booktabs}
\usepackage{tcolorbox}
\usepackage{amsmath}
\usepackage{subcaption}
\usepackage{pifont}  
\newcommand\torevise[1]{\textcolor{black}{#1}}
\usepackage{newfloat}
\usepackage{listings}
\DeclareCaptionStyle{ruled}{labelfont=normalfont,labelsep=colon,strut=off} % DO NOT CHANGE THIS
\floatstyle{ruled}
\newfloat{listing}{tb}{lst}{}
\floatname{listing}{Listing}
\title{Learning to Compress: \\Unlocking the Potential of Large Language Models  for Text Representation}
\author{
    Yeqin Zhang\textsuperscript{1,2},
    Yizheng Zhao\textsuperscript{1,2},
    Chen Hu\textsuperscript{3}, 
    Binxing Jiao\textsuperscript{3},
    Daxin Jiang\textsuperscript{3}, \\
    Ruihang Miao\textsuperscript{3}$^*$, 
    Cam-Tu Nguyen\textsuperscript{1,2}\thanks{Corresponding authors.}\\
}
\affiliations{
    \textsuperscript{1} State Key Laboratory for Novel Software Technology, Nanjing University, China\\
    \textsuperscript{2} School of Artificial Intelligence, Nanjing University, China\\
    \textsuperscript{3} Stepfun, China
    \\
     zhangyeqin@smail.nju.edu.cn, zhaoyz@nju.edu.cn\\
     \{hatcher, benjiao, djiang, miaoruihang\}@stepfun.com, 
     ncamtu@nju.edu.cn\\
}

\begin{document}
\maketitle
\begin{abstract}
 Text representation plays a critical role in tasks like clustering, retrieval, and other downstream applications. With the emergence of large language models (LLMs), there is increasing interest in harnessing their capabilities for this purpose. However, most of the LLMs are inherently causal and optimized for next-token prediction, making them suboptimal for producing holistic representations. To address this, recent studies introduced pretext tasks to adapt LLMs for text representation. Most of these tasks, however, rely on token-level prediction objectives, such as the masked next-token prediction (MNTP) used in LLM2Vec. In this work, we explore the untapped potential of context compression as a pretext task for unsupervised adaptation of LLMs. During compression pre-training, the model learns to generate compact memory tokens, which substitute the whole context for downstream sequence prediction. Experiments demonstrate that a well-designed compression objective can significantly enhance LLM-based text representations, outperforming models trained with token-level pretext tasks. Further improvements through contrastive learning produce a strong representation model (LLM2Comp) that outperforms contemporary LLM-based text encoders on a wide range of tasks while being more sample‑efficient, requiring significantly less training data. Code is available at https://github.com/longtaizi13579/LLM2Comp.
\end{abstract}

\section{Introduction}

\begin{figure*}[htbp]
    \centering
    \begin{subfigure}{0.30\linewidth}
        \centering
        \includegraphics[width=\linewidth]{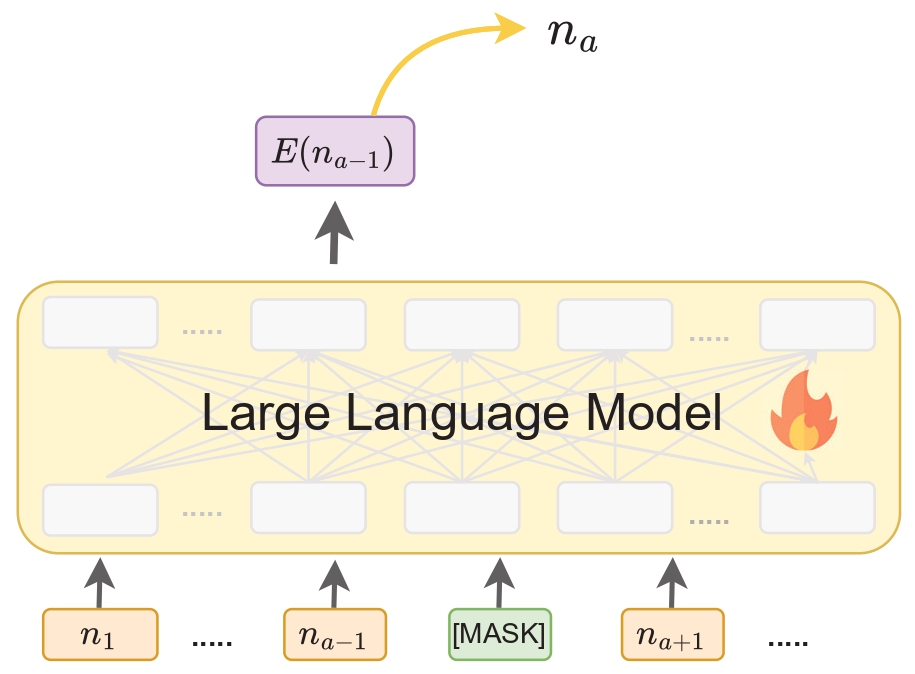}
        \caption{\llmtovec{}}
        \label{fig:llmtovec}
    \end{subfigure}
    \hspace{0.03\linewidth} % 调整间距
    \begin{subfigure}{0.30\linewidth}
        \centering
        \includegraphics[width=\linewidth]{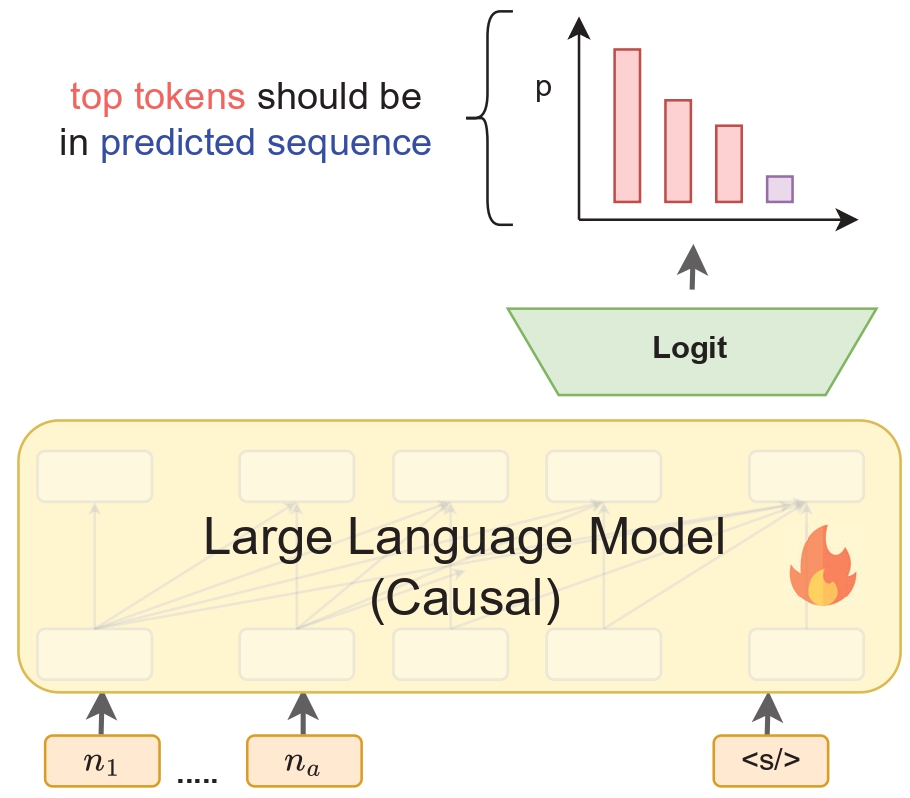}
        \caption{\lamatovec{}}
        \label{fig:lamatovec}
    
    \end{subfigure}
    \hspace{0.03\linewidth} % 调整间距
    \begin{subfigure}{0.30\linewidth}
        \centering
        \includegraphics[width=\linewidth]{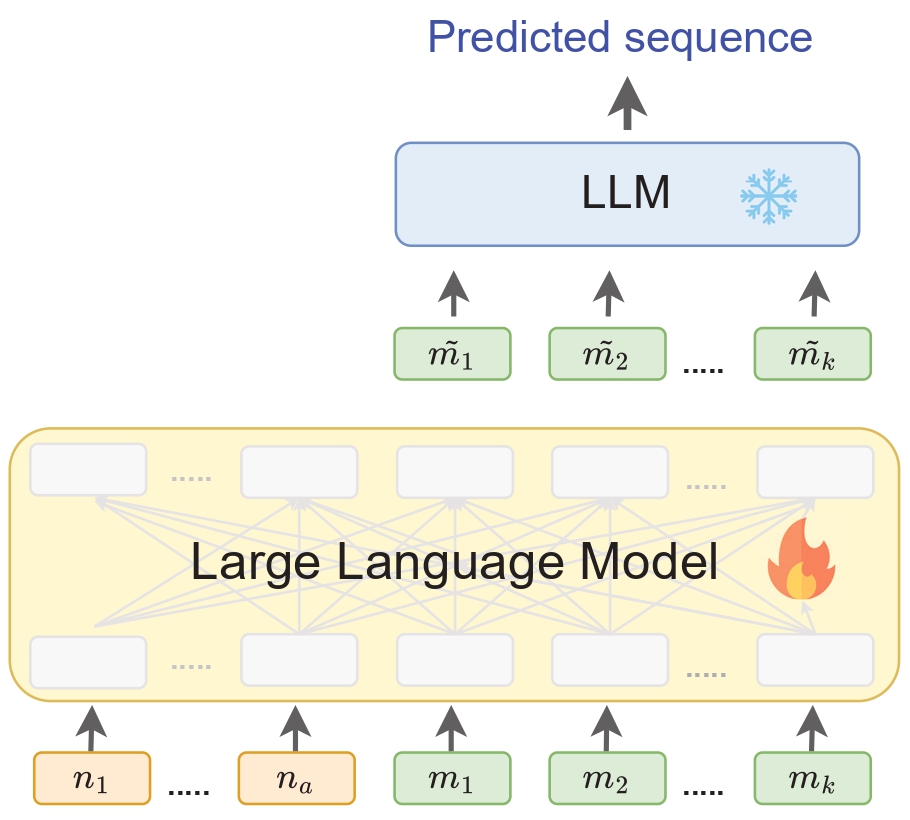}
        \caption{\ourmethod{}}
        \label{fig:ourmethod}
    \end{subfigure}
    \caption{The comparison of different pretext tasks: (1) MNTP (Mask Next Token Prediction) in \llmtovec{}; (2) EBAE or EBAR in \lamatovec{}; (3) Context Compression task in \ourmethod{} (ours)}
    \label{fig:meanpooling}
\end{figure*}
Text embedding models, which transform semantic content into vector representations, are fundamental to numerous tasks such as retrieval and recommendation. Early methods like TF-IDF and BM25 relied on simple statistics, thus falling short in capturing sequence semantics. The advent of deep neural networks led to the development of techniques such as Word2Vec \cite{DBLP:journals/corr/abs-1301-3781}. This paradigm shift paved the way for the foundational models like BERT \cite{DBLP:conf/naacl/DevlinCLT19} and T5 \cite{DBLP:journals/jmlr/RaffelSRLNMZLL20}. 

Recently, there has been growing interest in leveraging the powerful capabilities of LLMs for text representation. However, most LLMs are inherently causal and optimized for next-token prediction, which makes them inherently suboptimal for generating holistic, coherent representations of entire sequences. To address this limitation, recent studies have proposed various pretext tasks for unsupervised adaptation of LLMs, focusing primarily on token-level prediction objectives. For example, as shown in Figure~\ref{fig:llmtovec}, \llmtovec{} \cite{DBLP:journals/corr/abs-2404-05961} first transforms the causal attention mechanism of LLMs into a bidirectional form, and then adopts masked next token prediction (MNTP) as a pretext task to align the training objectives of causal LLMs with those of bidirectional models such as BERT. In this setup, MNTP randomly masks tokens within a sentence and leverages contextual representations of preceding tokens to predict masked ones. In contrast, as shown in Figure~\ref{fig:lamatovec}, \lamatovec{} \cite{DBLP:conf/acl/Li0XSL24} employs two pretext tasks, Embedding-Based Auto-Encoding (EBAE) and Embedding-Based Auto-Regression (EBAR), which predict tokens within the original sequence or the continued sequence. Although such objectives can capture the “bag-of-tokens” information, they cannot fully preserve the coherent semantic integrity of the entire sequence. These tasks, therefore, remain fundamentally token-level rather than sequence-level prediction.

In this work, we explore context compression (Figure~\ref{fig:meanpooling}) as a pretext task for the unsupervised adaptation of LLMs to text encoders. Specifically, during compression pre-training, the model learns to produce compact “memory tokens” that replace the original context for downstream sequence prediction tasks.  The conceptual difference between our objectives and previous pretext tasks is illustrated in Figure 1.  Here, we explore several compression objectives: \textit{reconstruction task} \cite{DBLP:conf/iclr/00010WWCW24, DBLP:conf/acl/XuFMHLWZLTZZC24,DBLP:conf/nips/0002W00CWZ024} involves regenerating original sentences;  \textit{continuation task} \cite{DBLP:conf/iclr/00010WWCW24, DBLP:conf/emnlp/ChevalierWAC23, DBLP:conf/acl/QinRCRD24, DBLP:conf/acl/XuFMHLWZLTZZC24, DBLP:journals/corr/abs-2401-07793} focuses on generating correct subsequent tokens. Our preliminary experiment shows that the reconstruction task does not provide satisfactory results, whereas the continuation task trained with NLL loss (CT-NLL) suffers from unstable training. Inspired by \cite{DBLP:conf/nips/Mu0G23, DBLP:conf/emnlp/WingateSS22}, we then propose \textit{Continuation Task with Knowledge Distillation} (CTKD) as a pretext task, which aims to predict the probability of subsequent tokens in alignment with the original extended sequence from the soft prompt. Our experiments demonstrate that this well-designed objective (CTKD) can significantly enhance LLM-based text representations.

We then performed a comprehensive analysis and found that LLMs' embeddings trained with the compression pretext task still suffer from dimensional collapse \cite{DBLP:conf/iclr/JingVLT22}, where the vectors lie in a low-dimensional subspace instead of using the full embedding space. However, compression pretraining with the CTKD objective is less affected by this issue than CT-NLL. This likely explains the advantage of CTKD over the CT-NLL objective. To further address dimensional collapse, we apply contrastive post-training on the model pretrained with CTKD. This post-training consists of an unsupervised contrastive phase followed by supervised contrastive learning (SCL). In particular, SCL encourages the embedding vectors of the negative samples to be pulled away, reducing the dimensional collapse and leading to significant performance gains. Our experiments show that CTKD pretraining synergizes with contrastive learning. As a result, our model, \ourmethod, outperforms other LLM-based models on many MTEB benchmark tasks. \ourmethod{} is also more sample-efficient, requiring much less (supervised) data during the contrastive learning phases than competing methods.

Our key contributions are summarized as follows:

\begin{itemize}
    \item We thoroughly investigate the untapped potential of compression pretraining for adapting LLMs to text representation tasks. Through empirical analysis, we provide insights into the crucial factors for the success of such a pretraining task, including the optimal training objective (CTKD) and the appropriate number of memory tokens.
\item We delve into the reasons behind the advantage of CTKD over other compression objectives, demonstrating that it is less prone to the dimensional collapse issue and thus more suited for downstream text representation.
\item We show that further enhancements through contrastive learning help alleviate the dimensional collapse issue. Note that the CTKD task provides a robust foundation for text representation, enabling our model, \ourmethod, to significantly outperform contemporary models such as \llmtovec{} and \lamatovec{} with less training data.
\end{itemize}

% \begin{figure*}[tbp]
% \centering
% \begin{subfigure}[b]{0.24\textwidth}
%     \centering
%     \includegraphics[width=\textwidth]{figure/reconstruction_new.pdf}
%     \caption{\scriptsize  Reconstruction }
%     \label{fig:reconstruction}
% \end{subfigure}
% %\hspace{0.2\textwidth} % Add horizontal space between subfigures
% \begin{subfigure}[b]{0.24\textwidth}
%     \centering
%     \includegraphics[width=\textwidth]{figure/continuation_nll_new.pdf}
%     \caption{\scriptsize  NLL Continuation}
%     \label{fig:continuation_nll}
% \end{subfigure}
% \begin{subfigure}[b]{0.48\textwidth}
%     \centering
%     \includegraphics[width=\textwidth]{figure/continuation_kl_new.pdf}
%     \caption{\scriptsize  KL Continuation}
%     \label{fig:continuation_kl}
% \end{subfigure}
% % \begin{subfigure}[b]{0.3\textwidth}
% %     \centering
% %     \includegraphics[width=\textwidth]{figure/qa.pdf}
% %     \caption{\scriptsize  Question Answering}
% %     \label{fig:qa}
% % \end{subfigure}
% \caption{\small Figure illustrating three unsupervised compression methods: (a) Reconstruction Task, (b) NLL Continuation Task, and (c) KL Continuation Task. \ncamtu{remove the text Large langauge model on top, make the fire and frozen sign to front, make the lines and text and icon of the causal LLM (frozen) inside the LLM box blur, make the borderline  of fire LLM box clear, show only KL}}
% \label{fig:three_compression_methods}
% \end{figure*}

\section{Related Works}
%\subsection{LLMs-based Text Representation}
Recent methods \cite{DBLP:journals/corr/abs-2412-09165} for adapting LLMs to text embedders can be broadly classified into training-free and training-based approaches.

\paragraph{Training-free Methods} The simple way to use LLMs as text encoders is to take the last token's hidden state of the final transformer layer as the representation. This is known as the last token pooling mechanism. However, the last token’s representation is optimized for next-token prediction rather than for aggregating global embedding. Recently, several prompting strategies such as PromptEOL \cite{DBLP:conf/emnlp/JiangHLWZ24} and MetaEOL \cite{DBLP:conf/acl/LeiW00CTY24} have been proposed to enhance the representational ability of the last token. An alternative approach is weighted mean pooling \cite{DBLP:journals/corr/abs-2202-08904}, which aggregates information from all sequence tokens. To further address the limitations of causal attention, EE \cite{DBLP:journals/corr/abs-2402-15449} duplicates the input text so that early tokens can attend to subsequent tokens. Although these training-free methods are simple, they still struggle to produce high-quality embeddings consistently. Moreover, approaches such as EE and MetaEOL increase the effective context length, which in turn increases the cost for extracting embeddings.

\paragraph{Training-required Methods} To better adapt LLMs for representation learning, a variety of training strategies have been proposed, including supervised contrastive learning \cite{DBLP:journals/corr/abs-2402-05672, DBLP:conf/sigir/MaWYWL24, DBLP:journals/corr/abs-2402-09906,DBLP:conf/iclr/LiQXCLLSL25,DBLP:conf/acl/WangYHYMW24, DBLP:journals/corr/abs-2412-03223,DBLP:journals/corr/abs-2402-09906,DBLP:conf/acl/Li0XSL24,DBLP:journals/corr/abs-2404-05961,su-etal-2025-training}, instruction tuning \cite{DBLP:conf/acl/SuSKWHOYSZ023}, and in-context learning \cite{DBLP:conf/iclr/LiQXCLLSL25}. Other research directions explore the use of synthetic data \cite{DBLP:conf/acl/WangYHYMW24, DBLP:journals/corr/abs-2412-03223}, multi-task learning (e.g., combining generation and representation learning) \cite{DBLP:journals/corr/abs-2402-09906,DBLP:journals/corr/abs-2408-03402}, variants of training loss functions \cite{deng-etal-2025-following}, or the use of pretext tasks \cite{DBLP:conf/acl/Li0XSL24,DBLP:journals/corr/abs-2404-05961}. Our work falls into the latter category, but focuses on the unique question of whether context compression can be utilized as an effective pretext task to enhance LLMs' ability to learn text representations.

\section{Text Representation via Compression}
\label{sec:rep-via-comp}
\subsection{Method}
For LLMs to capture global information from long contexts, we first convert them into bidirectional encoders that can process information in both directions. Building on this, we introduce context compression tasks as pretext tasks designed to further enhance their capacity to model the coherent semantics of entire contexts. Specifically, we consider two context compression tasks, reconstruction and continuation tasks \cite{DBLP:conf/iclr/00010WWCW24}, as described in detail below.

% Our training pipeline consists of three stages: (1) Compression, (2) Unsupervised Contrastive Learning, and (3) Supervised Contrastive Learning. In the following, we detail each stage.
% In the first stage, firstly, we transform the bi-directional attention to causal attention . We also train the compression task with causal attention to see the effect. Then we independently train three different compression models, each optimized using a distinct loss function. These models aim to condense the original input into a soft prompt that preserves essential semantic information. The three self-supervised objectives used are:

% \begin{figure}[tbp]
%     \centering
%     \includegraphics[width=0.45\textwidth]{figure/continuation_kl_new.pdf}
%     \caption{Continuation Task with Knowledge Distillation}
%     \label{fig:continuation_kl}
% % \caption{Figure illustrating three unsupervised compression methods: (a) Reconstruction Task, (b) NLL Continuation Task, and (c) KL Continuation Task. \ncamtu{remove the text Large langauge model on top, make the fire and frozen sign to front, make the lines and text and icon of the causal LLM (frozen) inside the LLM box blur, make the borderline  of fire LLM box clear, show only KL}}
% \label{fig:three_compression_methods}
% \end{figure}

\paragraph{Reconstruction Task} 
Let $g_\phi$ denote the original LLM, and $f_\theta$ be the targeted LLM-based (bidirectional) encoder adapted from $g_\phi$. The encoder $f_\theta$ is responsible for producing the embeddings of $k$ special (memory) tokens $\widetilde{m}_1, \ldots, \widetilde{m}_k$ given a long context $n_1, n_2, \ldots, n_a$. We say the set of compressed tokens $\{\widetilde{m}_i\}_{i=1}^k$ effectively captures the context if $g_\phi$ can reconstruct the context from them. This process is formalized as follows:
\begin{align}
    &\widetilde{m_i}= f_\theta(n_1,\ldots,n_a,m_1,\ldots,m_i)[-1]   \label{eq:f_theta}\\
    &\hat{n}_l=g_\phi(\widetilde{m_1}\ldots, \widetilde{m_k}, n_1, \ldots, n_{l-1})
\end{align}
\noindent where the memory embedding $\widetilde{m}_i$ for token $m_i$ is taken at the final hidden layer, just before the logit layer, as shown in Equation 1. Equation 2 describes how the original context is reconstructed using the memory embeddings. During training, we sample context $n_1,\ldots, n_a$ from a text collection, and train the encoder $f_\theta$ with LoRA \cite{DBLP:conf/iclr/HuSWALWWC22} while keeping the LLM $g_\phi$ frozen. Here, we use the negative log-likelihood computed by the frozen LLM 
$g_\phi$ to compare the original context with the reconstructed context, which is  generated from the compressed tokens.

\paragraph{Continuation Task} 
This variant is trained to predict future tokens given a compressed prefix.  Formally,
\begin{align}
    &\widetilde{m_i}= f_\theta(n_1,\ldots,n_a,m_1,\ldots,m_i)[-1]\\
    &\hat{n}_j=g_\phi(\widetilde{m_1}\ldots, \widetilde{m_k}, n_{a+1}, \ldots, n_{j-1})
\end{align}
where $(n_1, \ldots, n_a, n_{a+1}, \ldots, n_j)$ denotes a sentence from the dataset, 
split into two segments: a prefix $(n_1, \ldots, n_a)$ and a continuation $(n_{a+1}, \ldots, n_j)$. Similar to the reconstruction task, we adapt the encoder using LoRA and optimize it with the negative log-likelihood (NLL) computed by the frozen LLM $g_\phi$, which measures the discrepancy between the generated continuation and the ground-truth continuation. 

% We compress \(n_1, \ldots, n_a\) into a soft prompt and then task the model with generating \(n_{a+1}, \ldots, n_j\). The tokens \(m_1, \ldots, m_k\) are appended special tokens that gather the sentence's semantic information. The function \(f_\theta\) is the frozen language model with trainable LoRA parameters, which outputs the final hidden state representation. We only consider \(n_j\) as the ground truth token and optimize only its probability.

\paragraph{Continuation Task with Knowledge Distillation (CTKD)}
To further enhance encoder training, inspired by \cite{DBLP:conf/nips/Mu0G23, DBLP:conf/emnlp/WingateSS22}, we propose a third pretext task that combines the continuation objective with knowledge distillation. 
In this variant, the encoder $f_\theta$ is trained not only to generate the correct continuation, 
but also to match the next-token prediction distribution of the frozen LLM $g_\phi$ when conditioned on the compressed context versus the original context. This encourages distributional alignment between the two representations. Specifically, we build on the process of Equations 3 and 4, and exploit the Kullback-Leibler divergence loss  (KL loss) for training. The loss computation is formalized as follows:
\begin{align}
    % &KL (g_\phi(n_j \mid n_1,\ldots,n_{j-1}), g_\phi(n_j \mid \widetilde{m_1},\ldots,\widetilde{m_k})) \cdot \\
    &\quad\log\biggl(\frac{g_\phi(n_j \mid n_1,\ldots,n_a, n_{a+1}, \ldots,n_{j-1})}{g_\phi(n_j \mid \widetilde{m_1},\ldots,\widetilde{m_k},n_{a+1},\ldots,n_{j-1})}\biggr)
\end{align}

\paragraph{Mean Pooling}
Once the LLM has been adapted into the encoder $f_\theta$ using one of the proposed pretext tasks, the resulting encoder can be employed to generate text embeddings for a wide range of downstream tasks. In particular, a sentence embedding is obtained by applying mean pooling over the memory embeddings, as follows:
\begin{align}
    z = \frac{1}{k} \sum_{i=1}^{k} \widetilde{m_i}, \label{eq:embedding}
\end{align}

% \paragraph{Question Answering Loss.} The supervised model is trained to answer questions related to the input text using the compressed soft prompt. Formally,
% \begin{align}
%     \max_\theta &P(r_j \mid m_1, \ldots, m_k, q_1, \ldots, q_c, r_1, \ldots r_{j-1}),\\
%     &m_i = f_\theta(n_1,\ldots,n_b,\widetilde{m_1},\ldots,\widetilde{m_i})[-1],
% \end{align}
% where \(n_1, \ldots, n_b\) represents the long document or sentences, which are compressed into a soft prompt \(m_1, \ldots, m_k\). The tokens \(\widetilde{m_1}, \ldots, \widetilde{m_k}\) are appended special tokens that capture the sentence’s semantic information via self-attention. Based on the soft prompt and the question \(q_1, \ldots, q_c\), the model must generate the response \(r_1, \ldots, r_j\) accurately. The function \(f_\theta\) is the frozen language model with trainable LoRA parameters, which yields the final hidden state representation. In this task, we again consider all tokens \(r_j\) and optimize the distribution accordingly.

\begin{table*}[t]
\centering
\resizebox{\textwidth}{!}{
\begin{tabular}{lccccccccccccccccc}
\toprule
\multirow{2}{*}{\raisebox{-2ex}{\textbf{\centering Model}}} & \multirow{2}{*}{\raisebox{-2ex}{\makecell{\textbf{Training}\\\textbf{Samples}}}}& \multirow{2}{*}{\raisebox{-2ex}{\makecell{\textbf{Backbone}}}} & \multicolumn{3}{c}{Clustering} & \multicolumn{3}{c}{Retrieval} & \multicolumn{3}{c}{STS} & \multicolumn{3}{c}{Classification} & \multicolumn{2}{c}{Reranking} & \multicolumn{1}{c}{\multirow{2}{*}{\raisebox{-2ex}{Avg.}}} \\
\cmidrule(lr){4-6}\cmidrule(lr){7-9}\cmidrule(lr){10-12}\cmidrule(lr){13-15}\cmidrule(lr){16-17} 
 & & &\makecell{Bior.} & \makecell{Medr.} & \makecell{Twen.} & 
\makecell{SciF.} & \makecell{NFCo.} & \makecell{Argu.} & 
\makecell{STS17} & \makecell{SICK-R} & \makecell{STSB.} & 
\makecell{Bank.} & \makecell{Emot.} & 
\makecell{Spri.} & 
\makecell{Stac.} & \makecell{SciD.} &  \\
% \cmidrule(lr){2-4}\cmidrule(lr){5-7}\cmidrule(lr){8-10}\cmidrule(lr){11-12}\cmidrule(lr){13-13}\cmidrule(lr){14-15}
% & \makecell{75000}
\midrule
\multicolumn{16}{c}{\textbf{Training-free Methods \& Models trained with Pretext Tasks}}\\
\midrule
\textbf{LT.} & 0 & Llama-2& 15.99 & 17.42 & 15.96 & 2.17 & 1.31 & 14.24 & 57.8 & 55.63 & 45.72 & 68.65 & 29.85 & 47.01 & 32.07 & 58.83 & 33.05 \\  
\textbf{WMP.} & 0 & Llama-2 & 19.73 & 19.47 & 14.54 & 38.89 & 6.13 & \textbf{33.59} & 63.91 & 57.52 & 58.01 & 66.42 & 30.97 & 58.48 & 37.74 & 61.05 & 40.46 \\
\textbf{EE.} & 0 & Llama-2 & 22.94 & 23.15 & 25.74 & 25.61 & 9.97 & 25.24 & 80.51 & 70.18 & 71.94 & 81.79 & 45.00 & 68.48 & 40.79 & 60.15 & 46.54 \\
\textbf{PrompEOL} & 0 & Llama-2 & 22.49 & 21.14 &  31.47 & 27.16 & 13.59 & 11.65 & 79.67 & 73.82 & 75.32 & 76.37 & 47.13 & 26.08 & 37.65 & 66.22 & 43.55 \\
\textbf{MetaEOL} & 0 & Llama-2 & \textbf{30.95} & 26.56 & \textbf{40.03} & 40.59 & \textbf{16.41} & 21.75 & \textbf{82.29} & \textbf{76.88}& \textbf{76.87} & \textbf{82.26} & \textbf{51.05} & 48.24 & 39.87 & 77.91 & 50.83 \\
$\textbf{Llama2vec}$ & 32k & Llama-2 & 22.42 & 22.25 & 29.84 & 16.50 & 5.22 & 32.16 & 75.72 & 58.00 & 64.18 & 75.83 & 38.64 & 84.47 & 28.75 & 55.51 & 43.54\\
$\textbf{LLM2Vec}$ & 32k & Llama-2 & 26.44 & 25.14 & 25.76 & \textbf{44.51} & 4.34 & 31.02 & 73.45 & 67.65 & 65.82 & 79.77 & 39.28 & 70.07 & 41.48 & 61.48 & 46.87 \\
% $\textbf{Llm2Comp}_{causal}$ & 32k & 26.27 & 26.25 & 30.61 & 31.13 & 8.29 & 29.69 & 82.06 & 67.67 & 74.05 & 84.38 & 46.87 & 91.44 & 46.78 & 77.02 & 51.61 \\
 \textbf{\ourmethodrc} & 32k & Llama-2 & 6.65  & 13.56 & 8.94  & 17.41& 1.56  & 14.58  & 64.66 & 54.37 & 41.20 & 73.95 & 36.06 & 76.89 & 36.50 & 54.72 & 35.79 \\
\textbf{\ourmethodnll} & 32k & Llama-2 & 30.24  & \textbf{27.34 } & 37.25   & 11.93    & 3.55  & 24.69  & 70.65 & 64.57 &63.05 &80.12 &39.40  &72.02  &43.36 &78.94  & 46.22 \\
\textbf{\ourmethodkl} & 32k & Llama-2 & 27.79 &26.00  &31.19  &42.57  &9.24  &30.92  &81.56  &68.28&70.87  &84.33  &46.85 &\textbf{88.81 } &\textbf{48.20} & \textbf{78.18} & \textbf{52.49}\\
\bottomrule
\end{tabular}
}
\caption{Performance comparison of different models across three stages of training: self-supervised compression pretraining, evaluated on various tasks from the MTEB benchmark. Each model is assessed on a range of datasets, with results showing the impact of different training approaches on task performance.}
\label{tab:first_stage_result}
\end{table*}

\subsection{Experiment Setup}

\paragraph{Implementation Details} The unsupervised adaptation of LLMs using the proposed compression-based pretext tasks is 
conducted on \textbf{32,000 samples from the English Wikipedia-103 dataset} \cite{DBLP:conf/iclr/MerityX0S17}. This choice of dataset for unsupervised adaptation of LLMs to text embedders is consistent with the settings used in \textsc{LLM2Vec}  and \textsc{LLama2Vec}. This dataset is chosen because Wikipedia is included in the pretraining data of the LLM models, thus the adaptation process does not introduce new factual knowledge; rather, it focuses on teaching the model how to compress contextual information into soft tokens and construct sequence-level representations. For \ourmethod{}, we select the default number of memory tokens to be 8, unless stated otherwise. We provide additional details of our training setup and hyperparameters in the Appendix ~\ref{implement_details}.

\paragraph{Evaluation Datasets} \torevise{We evaluate the method across 14 diverse tasks in six categories, including clustering, retrieval, semantic textual similarity (STS), classification, and reranking. The semantic similarity tasks (SST) directly assess whether an embedding captures sentence-level semantics. These tasks cover various domains, such as biomedical text, scientific literature, software and programming, finance and banking, and customer support. Dataset details are provided in Table~\ref{tab:mteb-subset} in the Appendix~\ref{evaluation_detail}.}

\paragraph{Baseline Methods} We evaluate several established methods for sentence representation:

\begin{itemize}
    \item \textbf{LT} and \textbf{WMP} \cite{DBLP:journals/corr/abs-2202-08904} are training-free methods that obtain embedding from the last token (LT) or weighted mean pooling (WMP).
    \item \textbf{EE (Echo Embedding)} \cite{DBLP:journals/corr/abs-2402-15449}: Sentence representations are created by duplicating the sentence and applying mean pooling to the latter sentence's tokens.
    \item \textbf{PromptEOL} \cite{DBLP:conf/emnlp/JiangHLWZ24}: A prompt, “means in one word:”, is appended to the end of the sentence to enhance its representational capacity.
    \item \textbf{MetaEOL} \cite{DBLP:conf/acl/LeiW00CTY24} designs eight meta-task prompts with ChatGPT-4 to guide LLMs to form sentence representations from multiple perspectives.
    \item \textbf{\lamatovec{}} \cite{DBLP:conf/acl/Li0XSL24} and \textbf{\llmtovec{}} \cite{DBLP:journals/corr/abs-2404-05961} adapt LLMs for text representation with different pretext tasks, including EBAE, EBAR, and MNTP. The models are obtained by training LLama2 with pretext tasks using the same dataset as our method  (\ourmethod{}). Note that here, we consider \llmtovec{} and \lamatovec{} trained with pretext tasks, without subsequent contrastive learning.
    \item  \textbf{\ourmethod{}}: We compare three alternatives of our method, including \ourmethodrc{} that is based on the reconstruction task, \ourmethodnll{} that is trained with continuation task and NLL loss (CT-NLL), and \ourmethodkl{} which is trained with continuation task and knowledge distillation (CTKD).
    
\end{itemize}

\subsection{Experimental Results}
Table~\ref{tab:first_stage_result} presents the performance of training-free methods as well as unsupervised adaptation methods based on different pretext tasks. It is observable that the reconstruction objective offers only marginal gains: \ourmethodrc{} performs only slightly better than simple last-token pooling (LT). In contrast, continuation-based objectives lead to better improvements, making  \ourmethodnll{} matches the performance of \lamatovec{}. However, \textit{the CTKD objective proves to be more suitable than the CT-NLL objective for text representation}, leading to the superior performance of \ourmethodkl{} over \ourmethodnll{} and all other baselines. 

\begin{figure}
    \centering
    \includegraphics[width=0.95\linewidth, height=3cm]{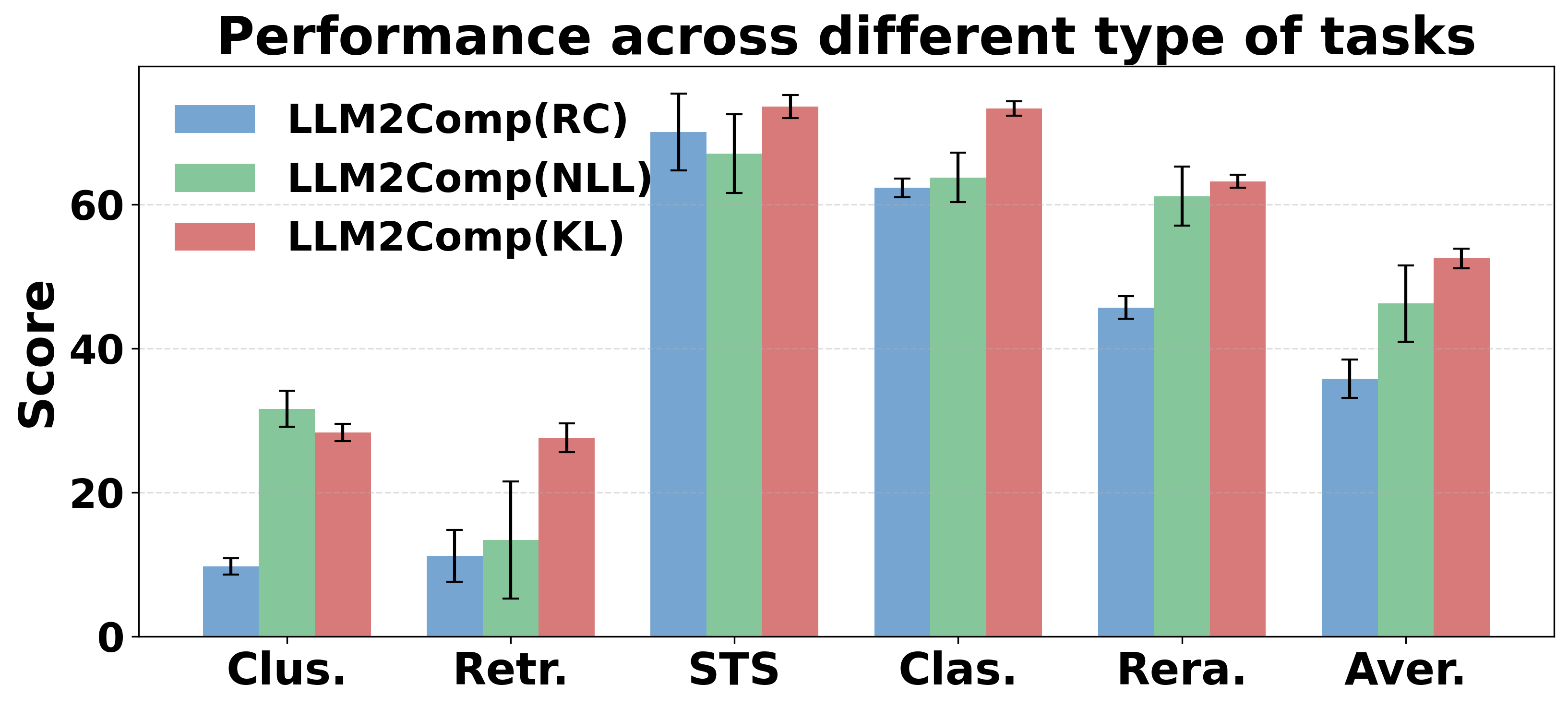}
    \caption{\torevise{Mean and standard variation of \ourmethodrc, \ourmethodnll, and \ourmethodkl across different task types, computed over five runs.}}
    \label{fig:first_stage_std}
\end{figure}

\paragraph{Stability of Different Compression Tasks} \torevise{In our experiments, we observe that \ourmethodkl{} exhibits more stable training behavior compared to \ourmethodnll{} and \ourmethodrc{}. As shown in Figure~\ref{fig:first_stage_std}, \ourmethodkl{} achieves a standard deviation of 1.37, which is significantly lower than 2.65 for \ourmethodrc{} and 5.32 for \ourmethodnll{}. Among the three, \ourmethodnll{} is the most unstable, as its performance can reach 51.85, comparable to \ourmethodkl{}, but in unfavorable cases, its performance drops to 42.95.}

\begin{figure}
    \centering
    \includegraphics[width=0.93\linewidth, height=3cm]{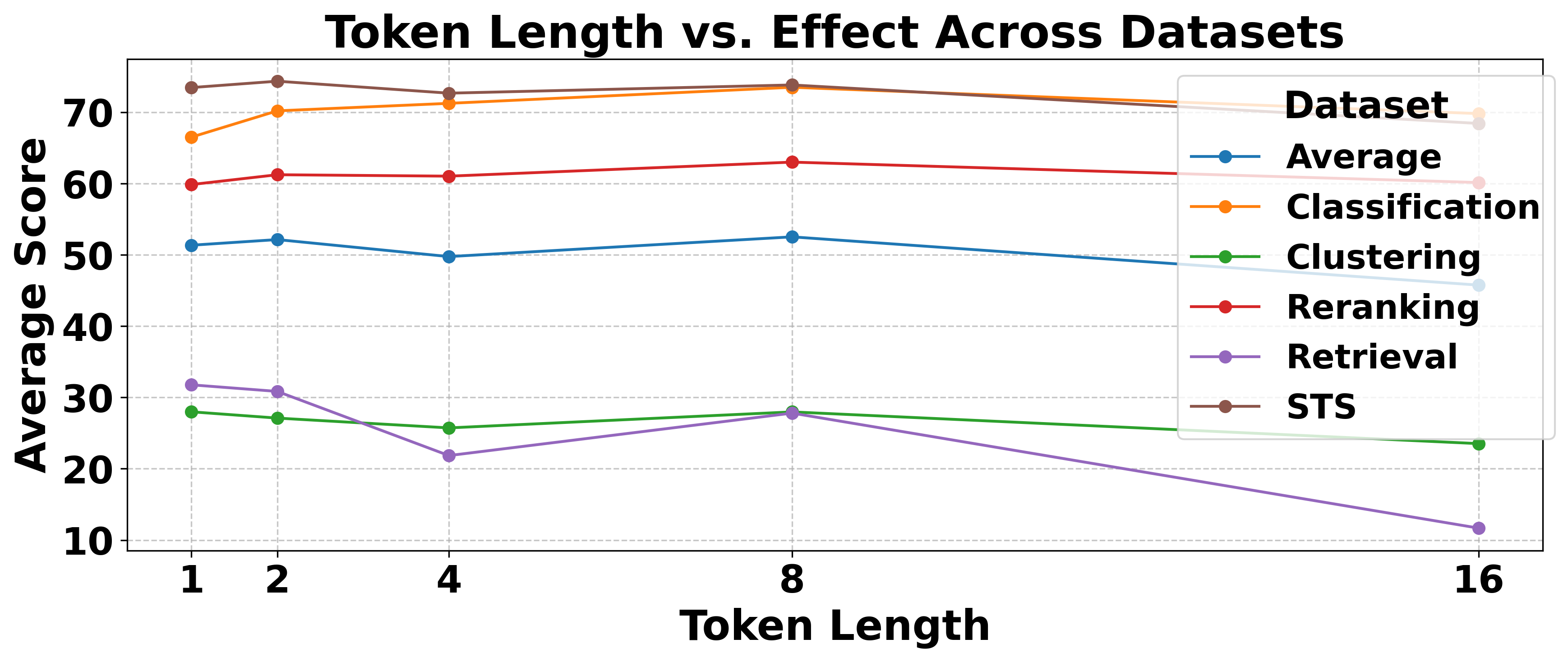}
    \caption{\ourmethodkl{}: Token length and its Effect}
    \label{fig:token_length_vs_effect}
\end{figure}

\label{token_length_impact}
\torevise{\paragraph{The Impact of Token Length}
Figure~\ref{fig:token_length_vs_effect} shows the impact of the memory token length on the performance of \ourmethodkl{}. When the number of tokens is in the range of [1, 8], performance remains stable across most tasks, except for retrieval, which is more sensitive to this hyperparameter. Specifically, when the number of tokens increases to 16, we observe a clear performance drop with the retrieval task. This observation contrasts with findings in context compression literature  \cite{DBLP:conf/acl/QinRCRD24, DBLP:conf/iclr/00010WWCW24, DBLP:conf/emnlp/WingateSS22}, where using a larger number of tokens (on the order of 100) facilitates  downstream generation.}

%Although previous research in compression learning \cite{DBLP:conf/acl/QinRCRD24, DBLP:conf/iclr/00010WWCW24, DBLP:conf/emnlp/WingateSS22} has demonstrated that increasing the number of compression tokens improves information capture and enhances downstream task performance, our experiments on token length yield a different conclusion under the representation learning setting. In the case of representation learning, as shown in Figure~\ref{fig:token_length_vs_effect}, the performance fluctuates as the token count increases from 1 to 8, suggesting a trade-off between information retention and the introduction of irrelevant dimensions. However, when the number of tokens increases from 8 to 16, a significant performance drop is observed, particularly in the retrieval task.
%\ncamtu{token length increase is good for compression but bad for compression-based representation with  mean pooling. Future work should focus on better pooling that is robust to the redundant tokens. The claim that incresing token length is good for compression can be taken from: https://arxiv.org/pdf/2210.03162}

\section{Dimensional Collapse}

%For Rc task, the compression tokens can directly see the ground truth, which makes this task simple. This simplicity may result in suboptimal generalization, as noted by \cite{DBLP:conf/iclr/00010WWCW24}. In this study, we investigate the factors that influence the performance of the NLL and KL continuation tasks. We need to note that, during several times training for NLL and KL continuation tasks, we find the evaluation result of KL is relatively stable, however, the result of NLL is very significant difference. Here we use the worst time of NLL to evaluate and analyse why it performs bad. We examine how these factors affect the generated compression tokens during inference. Our findings indicate that compression tokens derived from the NLL pretraining task exhibit greater internal similarity compared to those from the KL task. Moreover, under different random seeds, the results for NLL vary significantly, while those for KL remain relatively stable. Additionally, the performance of the NLL task is inferior when compared to that of the KL task. Building on these observations, we explore whether dimensional collapse affects the effectiveness of compression embeddings in both NLL and KL tasks. 

% \begin{figure}
%     \centering
%     \includegraphics[width=0.95\linewidth]{figure/time_and_effect.png}
%     \caption{Time consuming and Effect}
%     \label{fig:Effect_and_Time}
% \end{figure}

Since compression pretraining aims to condense long contexts into compact memory tokens, it may lead to dimensional collapse, a phenomenon also observed in self-supervised learning~\cite{DBLP:journals/entropy/ShwartzZivL24}. Here, dimensional collapse occurs when embedding vectors occupy a subspace of significantly lower dimensionality than the original embedding space. In our problem, this manifests in two ways: (i) the pooled embeddings exhibit low rank, and (ii) the resulting memory tokens become highly similar to one another. To study this effect, we analyze embeddings produced by \ourmethodkl{} and \ourmethodnll{} over a large corpus of 60K randomly sampled from the SciDocsRR dataset.
%Since our embeddings are obtained by mean-pooling the appended compression tokens, we analyze this issue from two perspectives: first, by examining the final embedding after mean-pooling across different sentences, and second, by analyzing the internal relationships between compression tokens based on embeddings derived from a large corpus of sentences.
\begin{figure}[tbp]
    \centering
    \includegraphics[width=0.95\linewidth, height=4cm]{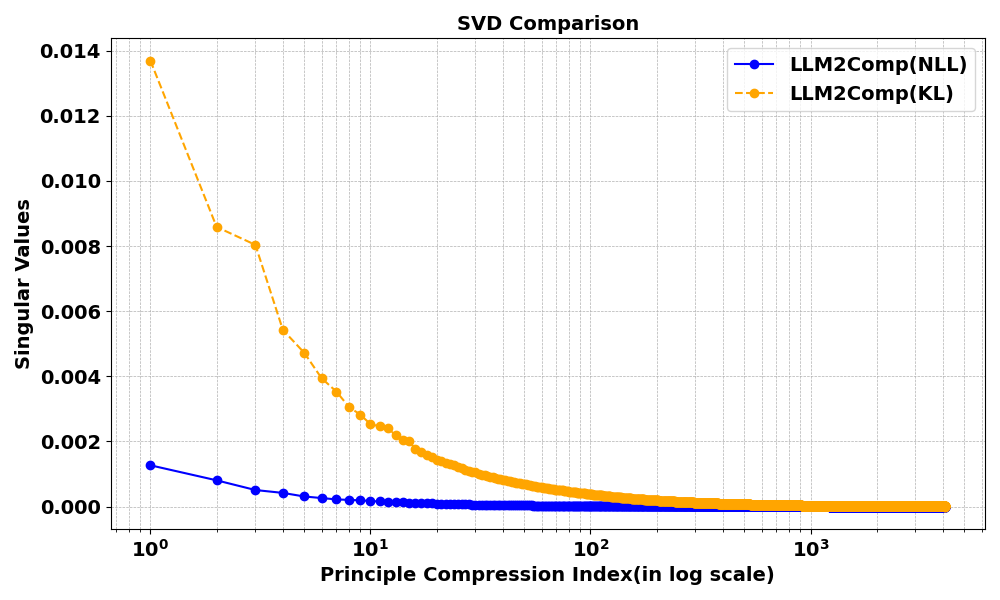}
    \caption{\torevise{Comparing singular values of \ourmethodnll{} and \ourmethodkl{}.}}
    \label{fig:pca_dim_first}
\end{figure}
\subsection{Effective Dimension After Meanpooling} In dimensional collapse, a small number of effective dimensions are sufficient to represent the data, while the remaining dimensions can be expressed as linear combinations of these effective ones. The degree of dimensional collapse can thus be quantified using Singular Value Decomposition (SVD)~\cite{DBLP:conf/iclr/JingVLT22}, where dimensions associated with near-zero singular values are considered ineffective. To this end, we first construct an embedding matrix from $N_s = 60{,}000$ samples, with each sample represented by a single embedding vector as defined in Equation~\ref{eq:embedding}. We then compute the covariance matrix of these embeddings, apply SVD, and sort the resulting singular values in descending order. The sorted order corresponds to the principal component index, where the first component corresponds to the largest singular value. We plot the resulting singular values against their principal component indices for \ourmethodkl{} and \ourmethodnll{} in Figure~\ref{fig:pca_dim_first}. Figure~\ref{fig:pca_dim_first} shows that the curve corresponding to \ourmethodnll{} drops to zero much faster than that of \ourmethodkl{}. The number of effective dimensions for \ourmethodnll{} and \ourmethodkl{} are in the order of 10 and 100, respectively. This result indicates that \ourmethodnll{} suffers more severe dimensional collapse in this case. \torevise{Intuitively, the KL divergence acts as a regularizer that better preserves information from less frequent tokens, thereby mitigating dimensional collapse. Nevertheless, the effective dimensionality of \ourmethodkl{} remains small relative to the total dimension (4096), indicating that there is still room for improvement.}

\subsection{Effective Token Index Before Meanpooling}
% Point here are: 
% 1. For NLL and KL, the bad performance of NLL is because existing many high similarity tokens.
% 2.  For good and bad performance of NLL, the bad performance of NLL is because existing many high similarity tokens.
% 3. More samples help reduce the similarity of high similarity tokens, but have little impact on low similarity samples.

\begin{figure}[t]
\centering
\begin{minipage}{0.231\textwidth}
    \centering
    \includegraphics[width=\textwidth]{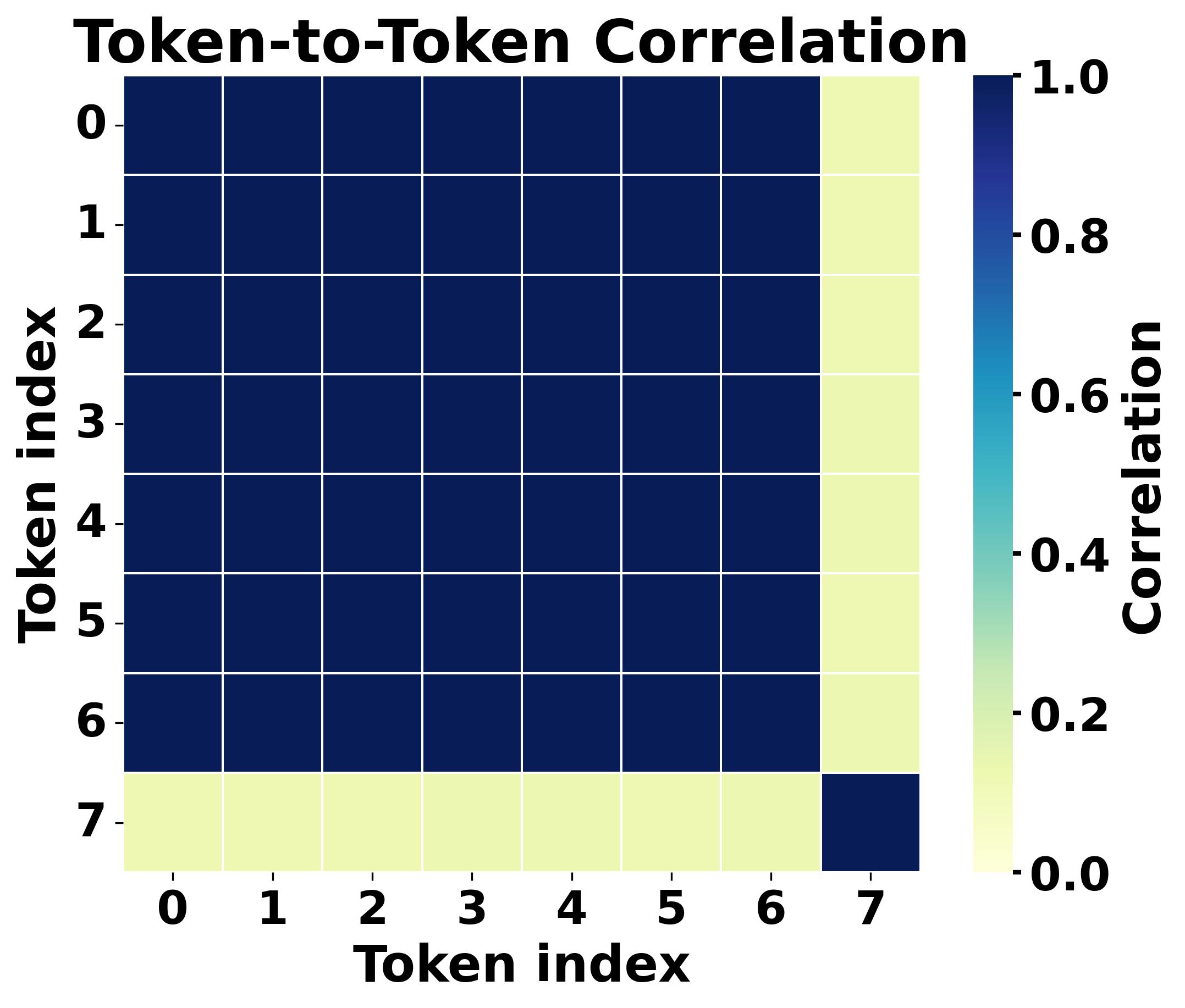}
    \label{fig:correlation_NLL_32000}
\end{minipage}
\hfill
\begin{minipage}{0.231\textwidth}
    \centering
    \includegraphics[width=\textwidth]{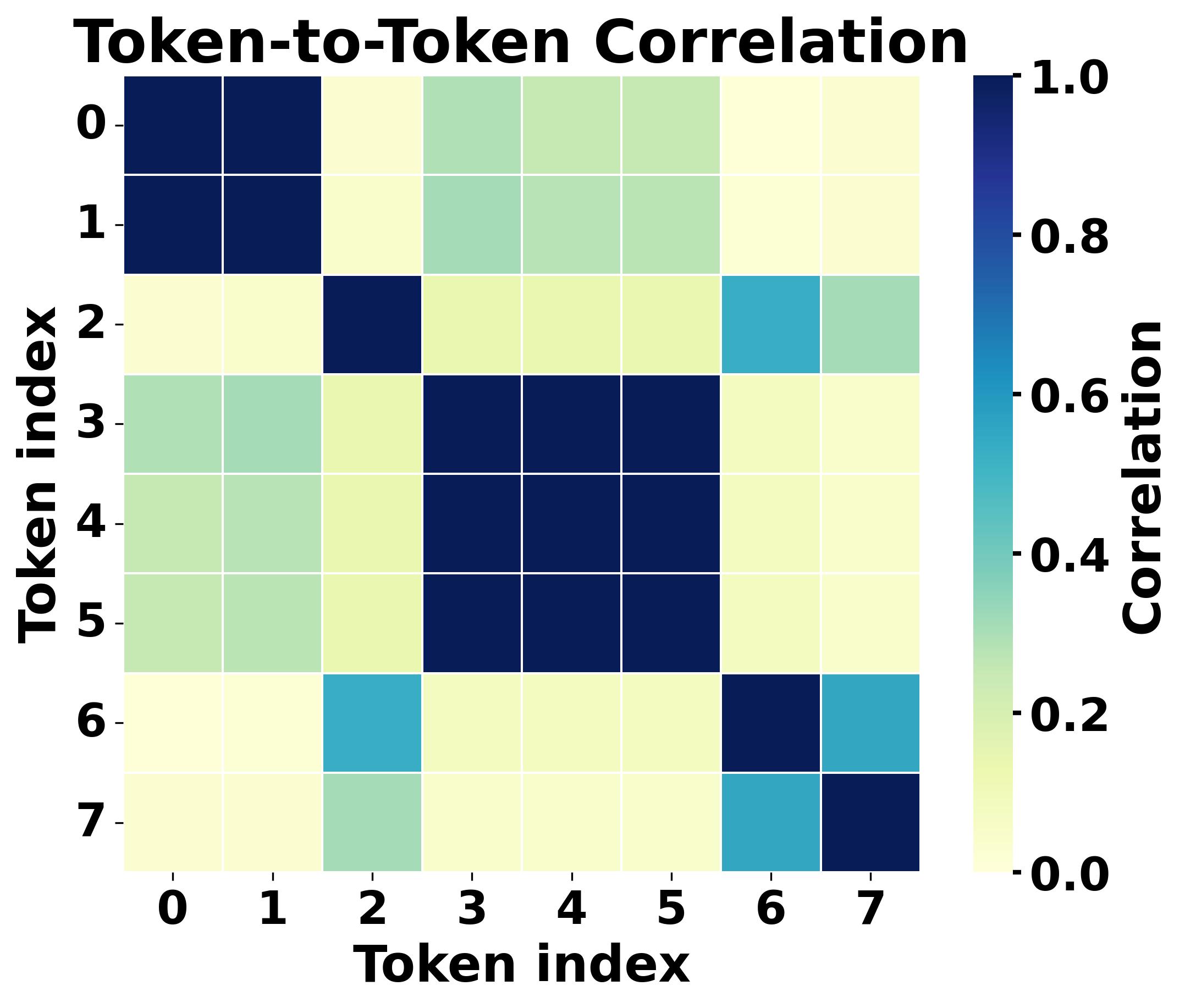}
    \label{fig:correlation_NLL_128000}
\end{minipage}
\caption{$\textbf{Llm2Comp}_{NLL}$ training with 32000 samples (left) and 128000 samples (right) in a bad training case.}
\label{fig:corrleation_nll}
\end{figure}

\begin{figure}[t]
\centering
\begin{minipage}{0.231\textwidth}
    \centering
    \includegraphics[width=\textwidth]{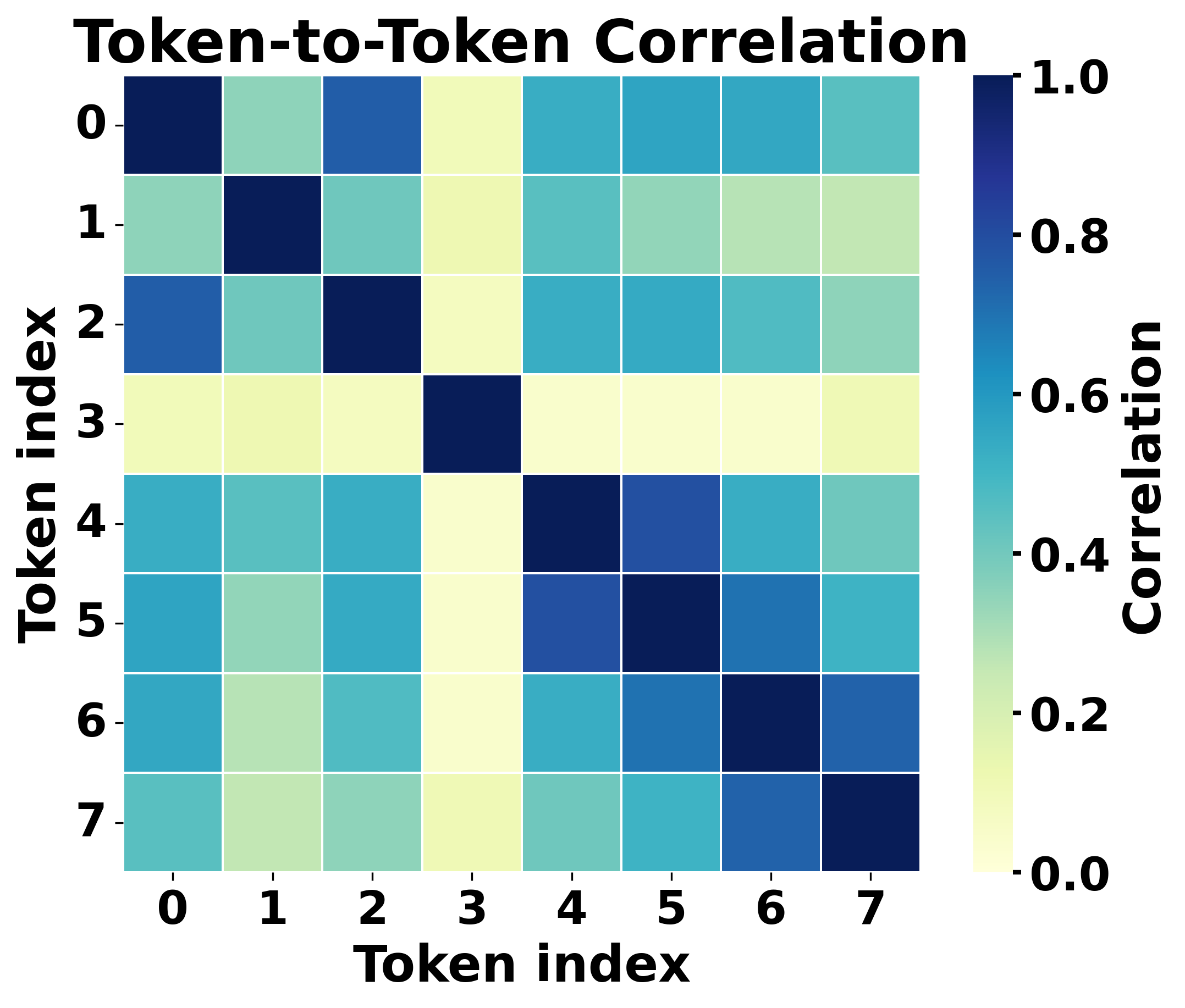}
    \label{fig:correlation_KL_32000}
\end{minipage}
\hfill
\begin{minipage}{0.231\textwidth}
    \centering
    \includegraphics[width=\textwidth]{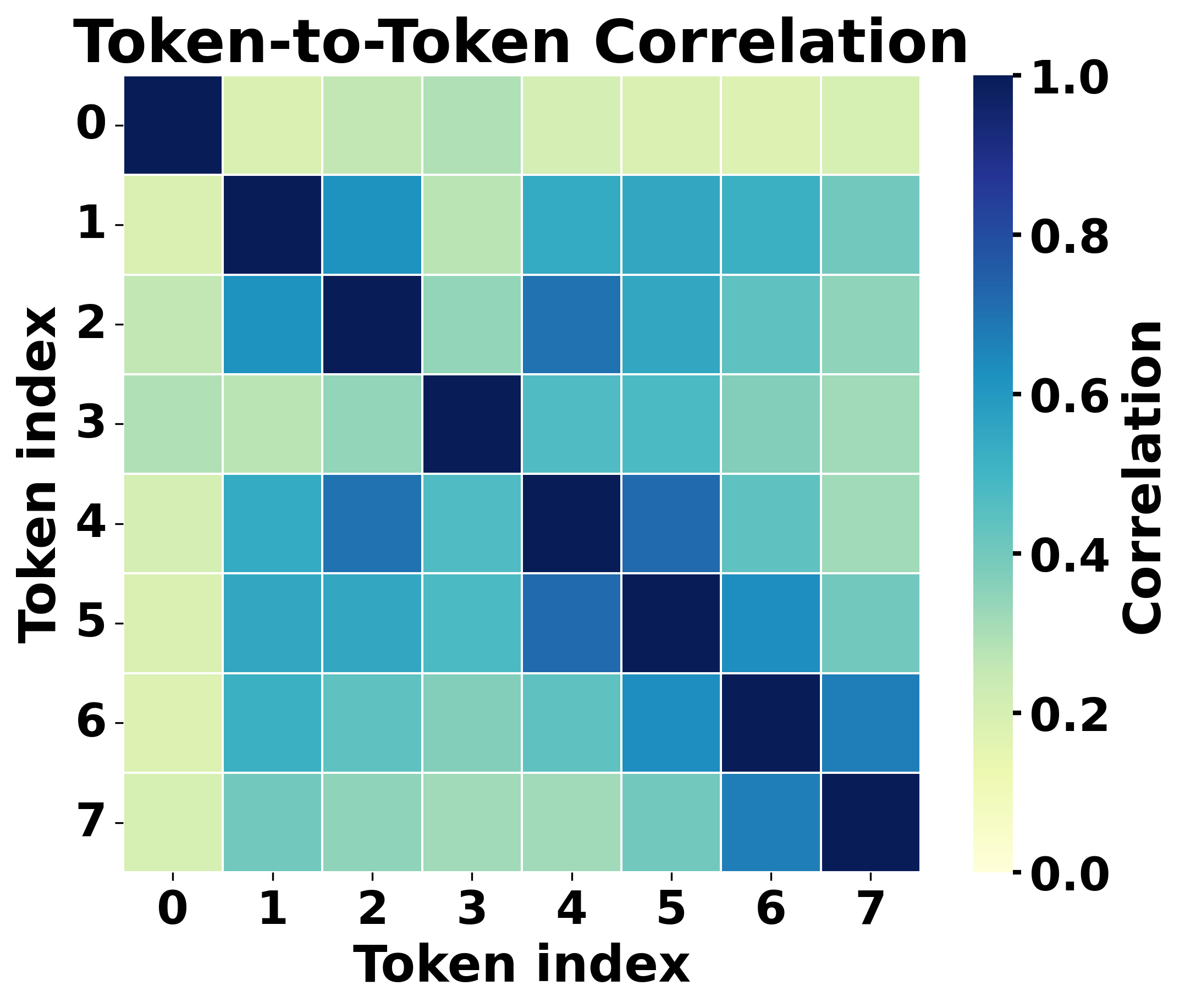}
    \label{fig:correlation_KL_128000}
\end{minipage}
\caption{$\textbf{Llm2Comp}_{KL}$ training with 32000 samples (left) and 128000 samples (right).}
\label{fig:correlation_KL}
\end{figure}

The dimensional collapse problem observed in models trained with compression objectives also manifests as a high degree of correlation among the memory tokens. To examine this, we compute the correlation matrix and average it over $N_s$ samples from ScidocRR as follows:
\begin{equation}
    Cor = \frac{1}{N_s}\sum_{i=1}^{N_s} \widetilde{M_i} \cdot \widetilde{M_i}^T
\end{equation}
Here, $\widetilde{M_i}$ denotes the matrix containing the compression tokens $\widetilde{m_{i1}}, \dots, \widetilde{m_{ik}}$ for the \( i \)-th sample. The correlation matrices of a typical \ourmethodnll{} model trained with \( N_s = 32\mathrm{K} \) and \( N_s = 128\mathrm{K} \) are presented in Figure~\ref{fig:corrleation_nll}. For comparison, Figure~\ref{fig:correlation_KL} shows the corresponding correlation matrices for a representative \ourmethodkl{} model trained with the same sample sizes.

Figure~\ref{fig:corrleation_nll} shows that tokens from \ourmethodnll{} are highly similar with 32K training samples. In contrast, \ourmethodkl{} with 32K samples suffers less from this issue. This observation is consistent with the analysis of effective dimensionality in the previous section. When the number of training samples increases to 128K, the token correlation problem in \ourmethodnll{} is partially alleviated. This improvement is also reflected in its performance, which rises from 42.95 to 48.38 as the sample size increases from 32K to 128K. These results suggest that the degree of token similarity also has a significant impact on the downstream task.

\begin{figure}
    \centering    
    \includegraphics[width=0.95\linewidth]{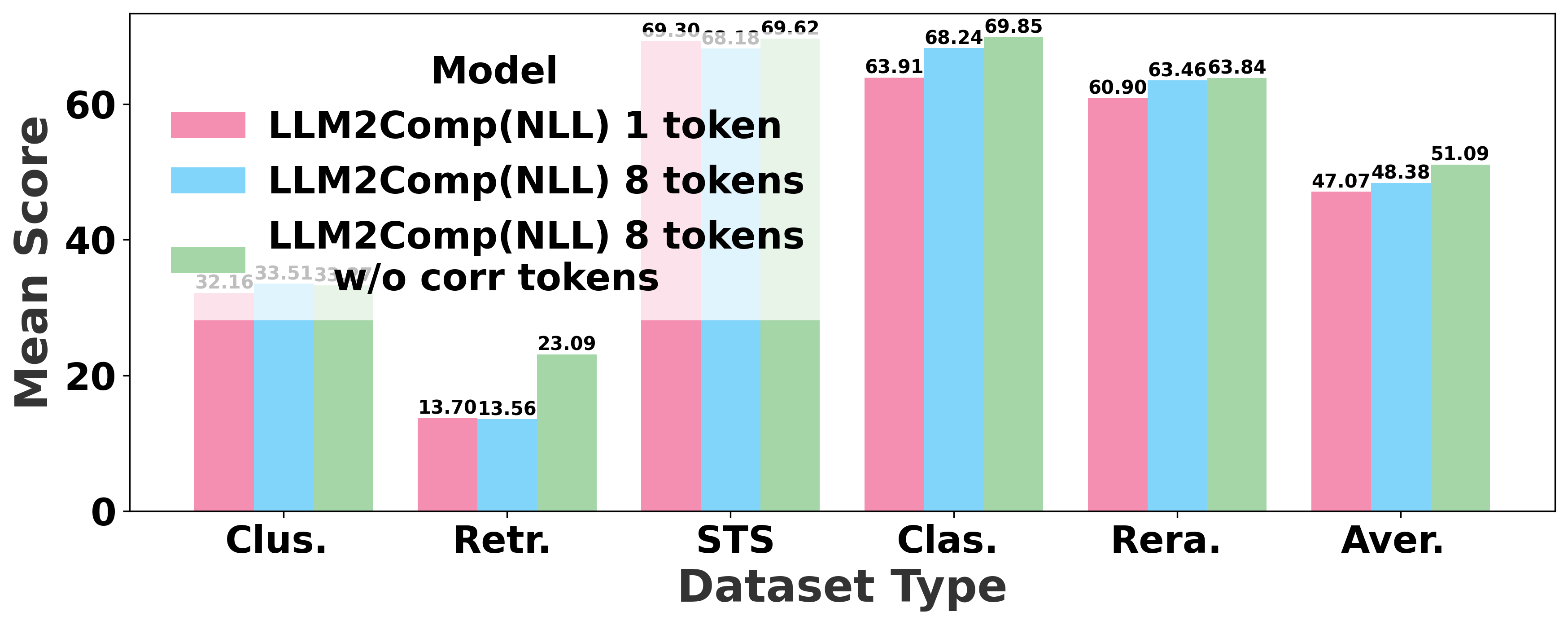}
    \caption{\torevise{Comparison between \ourmethodnll{} with 1-token compression and 8-token compression (with and without redundant tokens).}}
    \label{fig:effective_tokens_NLL}
\end{figure}

To further verify the above hypothesis, we investigate how reducing correlated tokens impacts downstream performance. We define a \emph{token cluster} as a set of tokens with high pairwise similarity. Initially, each token is treated as its cluster. Clusters are then merged if the minimum similarity between any pair of tokens across clusters exceeds 0.9. From each resulting cluster, we randomly select one representative token, and refer to these selected tokens as \emph{effective tokens}. Sentence embeddings are then computed by applying mean pooling to the embeddings of these effective tokens. Using this procedure, the performance of \ourmethodnll{} trained on 128,000 samples increases from 48.38 to 51.09, as shown in Figure~\ref{fig:effective_tokens_NLL}.  \torevise{We also include the performance of \ourmethodkl{} using a single compression token, as shown in Figure~\ref{fig:effective_tokens_NLL}. The results indicate that the single-token setting performs unsatisfactorily, likely due to excessive information loss. This highlights the need for a more effective learning strategy that better balances information preservation and redundancy}.

\section{Post-training with Contrastive Learning} \label{sec:cl-post-training}
A good representation should satisfy two key properties~\cite{DBLP:conf/icml/0001I20, DBLP:conf/iclr/JingVLT22}: (1) \emph{Alignment}, which encourages the representations of semantically related texts to be close to each other; and (2) a \emph{High Effective Dimensionality}~\cite{DBLP:conf/iclr/JingVLT22}, i.e., embedding vectors occupy much of the embedding space. While compression-based objectives implicitly promote alignment, they also tend to suffer from dimensional collapse, as shown in the previous section. Prior work~\cite{DBLP:conf/iclr/JingVLT22} has demonstrated that contrastive learning mitigates collapse by pushing representations of negative samples apart. In this paper, we investigate whether post-training with unsupervised contrastive learning (UCL) followed by supervised contrastive learning (SCL) can alleviate dimensional collapse and study its impact on downstream representation.

\subsection{Method}

\begin{table*}[t]
\centering
\resizebox{\textwidth}{!}{
\begin{tabular}{lccccccccccccccccc}
\toprule
\multirow{2}{*}{\raisebox{-2ex}{\textbf{\centering Model}}} & \multirow{2}{*}{\raisebox{-2ex}{\makecell{\textbf{Training}\\\textbf{Samples}}}}& \multirow{2}{*}{\raisebox{-2ex}{\makecell{\textbf{Backbone}}}} & \multicolumn{3}{c}{Clustering} & \multicolumn{3}{c}{Retrieval} & \multicolumn{3}{c}{STS} & \multicolumn{3}{c}{Classification} & \multicolumn{2}{c}{Reranking} & \multicolumn{1}{c}{\multirow{2}{*}{\raisebox{-2ex}{Avg.}}} \\
\cmidrule(lr){4-6}\cmidrule(lr){7-9}\cmidrule(lr){10-12}\cmidrule(lr){13-15}\cmidrule(lr){16-17} 
 & & &\makecell{Bior.} & \makecell{Medr.} & \makecell{Twen.} & 
\makecell{SciF.} & \makecell{NFCo.} & \makecell{Argu.} & 
\makecell{STS17} & \makecell{SICK-R} & \makecell{STSB.} & 
\makecell{Bank.} & \makecell{Emot.} & 
\makecell{Spri.} & 
\makecell{Stac.} & \makecell{SciD.} &  \\
% \cmidrule(lr){2-4}\cmidrule(lr){5-7}\cmidrule(lr){8-10}\cmidrule(lr){11-12}\cmidrule(lr){13-13}\cmidrule(lr){14-15}
% & \makecell{75000}
\midrule
\multicolumn{16}{c}{\textbf{Unsupervised contrastive learning (UCL)}}\\
\midrule
$\textbf{LLM2Vec}$ & 160k & Llama-2 & 31.25 & 28.04 & 30.76 & 64.48 & 26.81 & 47.09 & 86.70 & 71.77 & 78.32 & 84.65 & 46.58 & 87.57 & 47.77 & 77.62 & 57.82 \\
% $\textbf{Llm2Comp}_{causal}$ & 160k & 32.77 & 28.32 & 33.64 & 59.65& 30.91 & 31.78 & 87.27 & 73.69 & 79.58&86.32  &  48.56&  94.15& 51.50 & 80.94&  58.51\\
$\textbf{LLM2Comp}_{KL}$ & 160k & Llama-2 & 32.77 & 28.32 & 33.64 & 59.65& 30.91 & 31.78 & 87.27 & 73.69 & 79.58&86.32  &  48.56&  94.15& 51.50 & 80.94&  58.51\\
$\textbf{LLM2Comp}_{RC}$ & 160k & Llama-2 & 7.88 & 14.97 & 15.34 & 52.96 & 17.16 &25.05 & 76.55 & 58.52 & 60.32 & 75.77 &32.55 & 90.81 & 42.53 & 65.39 & 45.41\\
$\textbf{LLM2Comp}_{NLL}$ & 160k & Llama-2 & 31.03 & 26.65 & 35.97 & 55.49 & 27.58 & 27.47 & 86.61 & 75.43 & 77.69 &85.85 & 44.78 & 91.03 & 50.95 & 79.22 & 56.84\\
% $\textbf{Llm2Comp}_{NLL}$ & 160k  & 31.03 & 26.65 & 35.97 & 55.49& 27.58 & 27.47 &  86.61& 75.43 & 77.69 & 85.85 &  44.78& 91.03 &  50.95 & 79.22& 56.84 \\
% $\textbf{Llm2Comp}_{QA}$ & 160k & 23.46 & 21.17 & 28.34 & 64.15 & 21.64 & 43.05 & 84.43 & 65.91 & 78.84 & 84.20 & 51.17 & 93.33 & 47.77 & 74.49 & 55.85 \\
% $\textbf{Llm2Comp}_{Rc}$ & 160k & 5.56 & 11.96 & 12.61 & 33.51 & 6.64 & 23.78 & 74.59 & 58.26 & 59.92 & 76.06 & 35.23 & 89.41 & 41.85 & 59.70 & 42.08 \\
\midrule
\multicolumn{16}{c}{\textbf{Supervised contrastive learning (SCL)}}\\
\midrule
$\textbf{Instructor}$ & 1.4M & GTR-XL & 30.60 & 30.80 & 53.30 & 64.60 & 36.00 & 55.70 & 90.50 & 81.70 & 86.60 & 82.70 & 53.20 & 94.90 & 52.50 & 79.50 & 63.76 \\
% $\textbf{mE5}$ & ~1B & xlm-roberta-large & 33.50 & 29.70 & 38.90 & 70.40 & 34.00 & 54.40 & 88.10 & 80.20 & 87.30 & 84.70 & 46.50 & 93.10 & 49.70 & 82.00 & 62.32\\
% $\textbf{mE5}_{Instruct}$ & ~1B & xlm-roberta-large & 36.30 & 35.50 & 51.30 & 71.80 & 35.50 & 58.40 & 90.00 & 81.70 & 88.40 & 85.70 & 51.50 & 91.20 & 51.50 & 85.90 & 65.34\\
$\textbf{ULLME}$  & 0.5 M & Phi-1.5 & 30.46 & 30.18 & 42.95 & 63.41 & 34.54 & 55.06 & 88.49 & 70.49 & 80.81 & 84.24 & 45.83 & 92.78 & 48.61 & 79.29 & 60.51\\
$\textbf{ULLME}$ & 0.5 M & Mistral-v0.2 & 31.48 & 26.95 & 38.52 & 72.86 & 39.37 & 45.93 & 86.38 & 70.31 & 78.21 & 84.57 & 45.02 & 92.20 & \textbf{52.56} & 83.47 & 60.56\\
$\textbf{ULLME}$ & 0.5 M & Llama-3 & 30.32 & 26.01 & 41.32 & 72.38 & 39.37 & 46.78 & 86.30 & 69.11 & 80.25 & 84.76 & 49.48 & 94.73 & 52.38 & 81.42 & 61.05\\
% $\textbf{MLTP}$ (model-1) & 1.4 M & & 36.86 & 32.48 & 49.31 & 75.82 & 36.09 & 48.63 & 82.69 & 86.59 & --   & 48.71 & --   & --   & --   & 82.55 & --\\
% $\textbf{MLTP}$ (model-2) & 1.4 M & &37.56 & 33.39 & 46.61 & 73.32 & 37.48 & 54.08 & 83.19 & 83.09 & --   & 46.10 & --   & --   & --   & 83.09 & --\\
% $\textbf{MLTP}$ (model-3) & 1.4 M & &37.65 & 34.36 & 47.18 & 74.74 & 36.99 & 50.52 & 83.53 & 83.16 & --   & 48.75 & --   & --   & --   & 83.16 & --\\
% $\textbf{MLTP}$ (model-4) & 1.4 M & &32.89 & 30.68 & 43.29 & 74.18 & 39.01 & 54.89 & 83.45 & 83.92 & --   & 46.38 & --   & --   & --   & 83.92 & --\\
% $\textbf{MLTP}$ (model-5) & 1.4 M & &35.34 & 32.91 & 44.55 & 74.03 & 38.08 & 55.51 & 83.04 & 83.89 & --   & 50.37 & --   & --   & --   & 83.89 & --\\

%$\textbf{MLTP}_{bi-multilayer}$ & 1.4 M & & \\
\textbf{BGE-ICL} & 2 M & Mistral-v0.1 &35.00 & 28.10 & 43.65 & \textbf{78.10} & 40.16 & 55.81 & \textbf{91.65} & \textbf{83.83} & 87.27 & 87.57 & \textbf{54.29} & 94.79 & 51.48 & 84.31 & 65.43\\
\textbf{RepLlama} & 0.5M & Llama-2& - & -& - & 75.60 & 37.80 & 45.60 & - & - &  -&-  &-  & - & - & -&  -\\
\textbf{Llama2vec} & $>$3M &Llama-2& 30.38 & 28.21 & 45.63 & 75.95 & 37.38 & 49.08 & 66.73 & 68.57 & 71.61 & 77.05 & 46.17 & 95.65 & 45.87 & 77.04 & 58.24 \\
$\textbf{LLM2Vec}$  & 1.16M & Llama-2 &34.81 & 31.37 & 51.04 & 77.30 & \textbf{40.33} & 56.53 & 90.63 & 83.01 & \textbf{88.72} & \textbf{88.17} & 51.71 & \textbf{96.83} & 51.02 & 84.03 & 66.11 \\
%$\textbf{LLM2Vec}$  & 1.16M & Mistral-v0.2 &35.53 & 31.27 & 52.18 & 78.86 & 39.33 & 57.48 & 90.19 & 83.70 & 88.65 & 88.31 & 52.05 & 96.82 & 54.41 & 83.80 & 66.61 \\
\midrule
% $\textbf{Llm2Comp}_{causal}$ & 160k & 32.77 & 28.32 & 33.64 & 59.65& 30.91 & 31.78 & 87.27 & 73.69 & 79.58&86.32  &  48.56&  94.15& 51.50 & 80.94&  58.51\\
$\textbf{LLM2Comp}_{KL}$ & 0.36M & Llama-2  & \textbf{37.15} & \textbf{33.70} & \textbf{55.11} & 76.79 & 39.72 & \textbf{59.37} & 91.40 & 83.32 & 86.31 & 84.57 & 54.01 & 96.27 & 51.90 & \textbf{85.36} & \textbf{66.78} \\
$\textbf{LLM2Comp}_{RC}$ & 0.36M & Llama-2 & 34.81 & 31.30 & 53.63 & 73.54 & 37.55 & 57.73 & 90.90&83.17 & 86.24 & 83.61 & 53.65 &95.93 & 51.91 & 82.87 & 65.49 \\
$\textbf{LLM2Comp}_{NLL}$ & 0.36M & Llama-2 & 36.53 &32.85 & 53.14 & 75.05 & 39.13 & 58.20 &91.61 &83.05 & 85.33 & 82.99&52.24 &96.16 & 51.45 &84.85 &65.90 \\
% $\textbf{Llm2Comp}_{NLL}^3$ & 39.29 & 31.88 & 52.43 & 73.42 & 33.21 & 39.29 & 91.71 & 82.86 & 88.80 & 87.72 & 51.18 & 96.10 & 49.94 & 78.20 & 64.00 \\
% $\textbf{Llm2Comp}_{QA}^3$ & 35.23 & 31.18 & 49.12 & 69.60 & 30.51 & 35.30 & 92.02 & 82.62 & 87.68 & 87.02 & 52.46 & 96.07 & 45.31 & 69.66 & 61.70 \\
% $\textbf{Llm2Comp}_{Rc}^3$ & 35.78 & 32.63 & 51.29 & 72.14 & 33.53 & 43.54 & 92.12 & 81.99 & 89.29 & 87.74 & 52.46 & 96.29 & 49.37 & 78.52 & 64.05 \\
\bottomrule
\end{tabular}
}
\caption{Performance comparison of different models across different post-training stages. Here, the backbone models include: Llama-2 (7B), Mistral-v0.1 (7B), GTR-XL (1.5B), Phi-1.5 (1.3B), Mistral-v0.2 (7B), and Llama-3 (8B).}
\label{tab:two_stage_results}
\end{table*}

\paragraph{Unsupervised Contrastive Learning} Following SimCSE \cite{DBLP:conf/emnlp/GaoYC21}, we construct \textit{positive samples of a particular sentence through dropout, and treat other sentence samples as negative ones}. Formally, the objective is to train the encoder $f_\theta$ to maximize the InfoNCE loss:
\begin{align}
    \max_\theta &\log \frac{\exp(\mathrm{sim}(z_i, z_j)/\tau)}{\sum_{k=1}^{2B}  \exp(\mathrm{sim}(z_i, z_k)/\tau)}
    \label{eq:ucl}
\end{align}
Where \( z_{i} \) and \( z_{j} \) are the representations of the sentences obtained by meanpooling the memory tokens sequences $x_i$ and $x_j$ using $f_\theta$ (see Equation \ref{eq:f_theta}, and Equation \ref{eq:embedding}). Additionally, \( \mathrm{sim}(\cdot, \cdot) \) denotes the cosine similarity, \( \tau \) is the temperature hyperparameter, and \( B \) is the batch size. For UCL, in-batch negative sampling is exploited, i.e., positive embedding of one sample in the batch is considered as the sampled negative for others in the batch.

\paragraph{Supervised Contrastive Learning}
Following UCL, we perform SCL based on supervised data, where relevant pairs are manually annotated. Negative samples are chosen following in-batch negative sampling and hard-negative sampling. The hard-negative samples are pre-chosen by the E5 dataset from a cross-encoder model \cite{DBLP:journals/corr/abs-2402-05672}. The optimization objective is similar to UCL (Equation \ref{eq:ucl}) except that we have manually labeled positive samples. 

% In the third stage, we incorporate supervised contrastive learning to explicitly leverage semantic labels. For each input, the goal is to maximize the similarity between each sample and its  and its positive samples while minimizing the similarity with negative samples within the batch. Formally, the objective is:
% \begin{align}
%     \max_\theta &\log \frac{\sum_{x^+ \in \mathcal{P}(x)} \exp(\mathrm{sim}(z, z^+)/\tau)}{\sum_{x' \in \mathcal{P}(x) \cup \mathcal{N}(x)} \exp(\mathrm{sim}(z, z')/\tau)}
% \end{align}
% where \( z = f_{\theta}(x) \) and \( z^+ = f_{\theta}(x^+) \) also obtained by the equation \ref{eq:embedding}, \( \mathcal{P}(x) \) represents the set of positive samples, and \( \mathcal{N}(x) \) represents the set of negative samples. In addition to the in-batch negative samples used like the unsupervised stage, hard negative samples are also sampled out. These samples are similar to the positive samples. By distinguishing between positive and hard negative samples, we can better address dimensional collapse and recover more effective dimensions.

\subsection{Experimental Setup}

\begin{figure}[tbp]
    \centering
  \resizebox{0.8\linewidth}{!}{
    \includegraphics{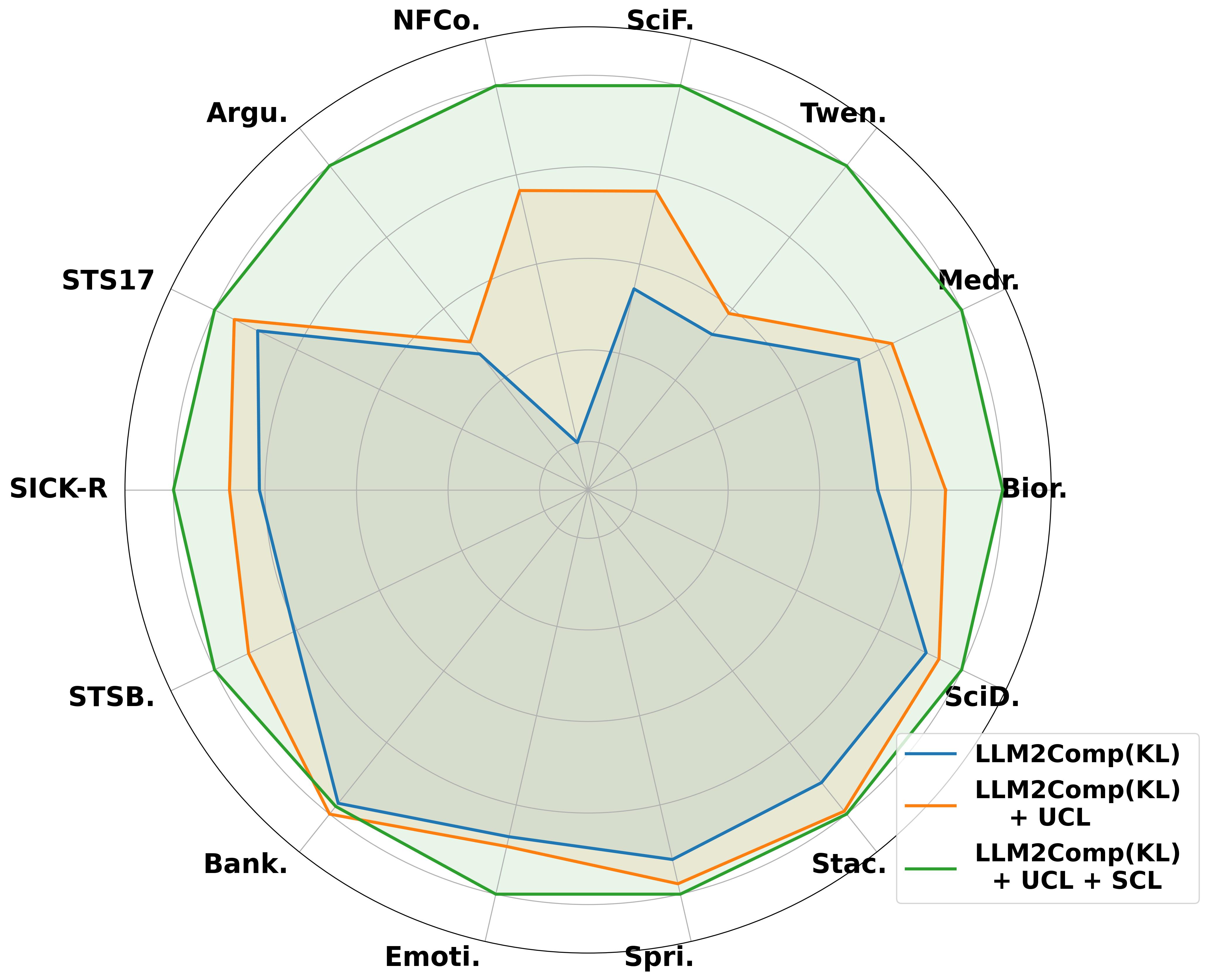}
  }
    \caption{Radar plot showing the performance of \ourmethodkl{} across different training stages and datasets.}
    \label{fig:different_stage_radar}
\end{figure}

\paragraph{Training Data} In the UCL stage, we utilize a Wikipedia sentence subset (128,000 samples) \cite{DBLP:conf/emnlp/GaoYC21} identical to LLM2Vec's second-stage training data to ensure comparable experimental conditions. In the SCL stage, we utilize 1,024,000 samples from the public portions of datasets employed in \llmtovec{}.

\paragraph{Compared methods}  In the following, unless otherwise specified, \ourmethod{} refers to the model built upon \ourmethodkl{} as the foundation model, further enhanced through contrastive post-training. In the UCL stage, we compare \ourmethod{} with \llmtovec{}, which is also first adapted using a pretext task (MNTP) and then trained with UCL. To maintain appropriate data variance, we apply a dropout rate of 0.2 for creating positive samples, which helps prevent excessive augmentation that could distort the original dataset distribution~\cite{DBLP:conf/iclr/JingVLT22}.

In the SCL stage, we further train \ourmethod{} initialized from the UCL stage. Our baselines include \llmtovec{}, \lamatovec{}, and RepLLaMA, all of which share the same LLM backbone and are trained with SCL. RepLLaMA~\cite{DBLP:conf/sigir/MaWYWL24} directly applies last-token pooling with SCL without using any pretext tasks. \lamatovec{} is trained with SCL directly after training with pretext tasks, as reported in the original paper \cite{DBLP:conf/acl/Li0XSL24}. We also report the performance of several contemporary LLM-based encoders, including ULLME, BGE-ICL (in a zero-shot setting), and Instructor. These models adopt different base models and different training strategies, such as finetuning with in-context data in BGE-ICL \cite{DBLP:conf/iclr/LiQXCLLSL25}, and multi-task learning in ULLME  \cite{DBLP:journals/corr/abs-2408-03402}. Consequently, comparisons with these models should be viewed as a reference only, as they do not directly provide scientific insight. Details of the compared methods are given in the Appendix~\ref{compared_methods}.

\begin{figure}[tbp]
    \centering
    \includegraphics[width=0.95\linewidth, height=3.7cm]{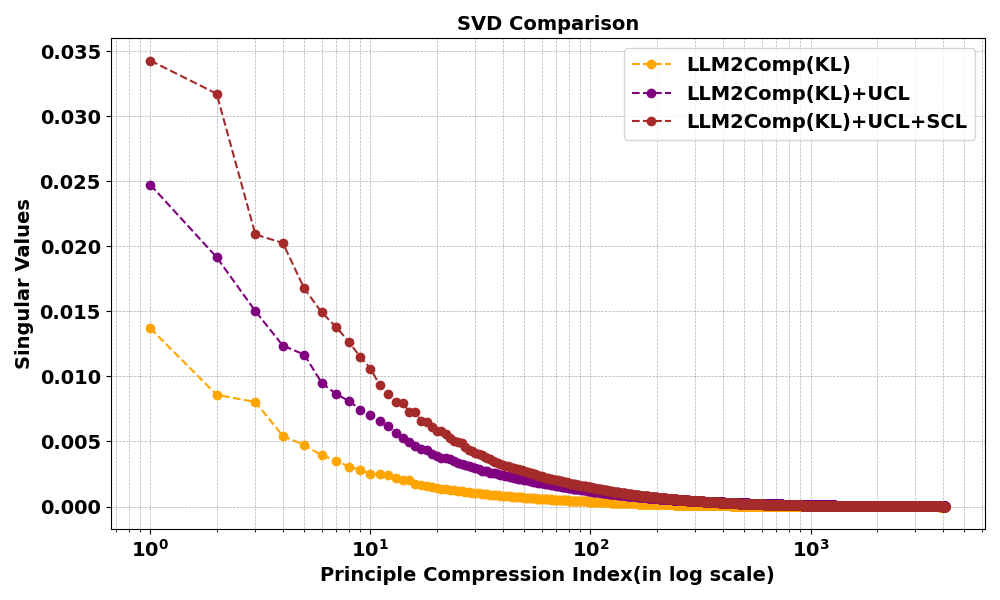}
    \caption{Comparing singular values of \ourmethodkl{} in different training stages.}
    \label{fig:pca_dim_second}
\end{figure}

It is worth noting that several of these models, including Instructor, BGE-ICL, and LLM2Vec, have reported results on MTEB, which includes our evaluation datasets. For these models, we directly use the scores reported in their respective papers. For others, such as Llama2Vec and ULLME, which only provide partial MTEB results, we perform our evaluations using their published models.

\subsection{Experimental Results}
Figure~\ref{fig:different_stage_radar} shows the performance of \ourmethod{} across different training stages and tasks. The results verify the contribution of UCL and SCL to performance gains beyond pre-training with pretext tasks. In addition, it is observable that the CL stages play a more important role in improving retrieval and clustering tasks.

Compared to other baselines, the experimental results in Table~\ref{tab:two_stage_results} show that \ourmethod{} achieves the best performance on most datasets as well as on average. In the UCL stage, \ourmethod{} surpasses \llmtovec{}, confirming that the benefits of our compression-based pretext task carry over to the subsequent training. A similar pattern is observed in the SCL stage after UCL, where \ourmethod{} is superior to \llmtovec{}, and other contemporary models. Notably, \ourmethod{} achieves this using a much smaller amount of supervised data compared to \llmtovec{} as shown in Table \ref{tab:two_stage_results}. This suggests that compression-based pretraining provides a stronger foundation, enabling more efficient post-training. Since training cost scales with the amount of supervised data, this also highlights the practical value of \ourmethod{}.

% Compression Pretraining unlocks the representation learning
% The results from the supervised contrastive learning stage are shown in the corresponding section of Figure~\ref{fig:different_stage_radar}. As observed, almost all datasets exhibit improved performance relative to Stage 2, with the exception of the Banking77Classification dataset, which shows a decline in performance. The retrieval tasks including ArguAna, NFCorpus, and SciFact demonstrated the most significant improvements. As anticipated, supervised learning, in contrast to unsupervised learning, necessitates the inclusion of documents that can directly respond to queries as positive samples, rather than enhancing positive samples through dropout techniques. Consequently, supervsied contrastive learning optimize the alignment (as dropout-based unsupervised learning mainly optimizes uniformity, with minimal emphasis on alignment). Additionally, through hard negative sampling and larger batch sizes, uniformity is further more optimized than unsupervised contrastive learning. As predicted, the results show that improved optimization of both alignment and uniformity leads to better overall performance.

\begin{figure}[tbp]
    \centering
    \includegraphics[width=0.95\linewidth, height=4cm]{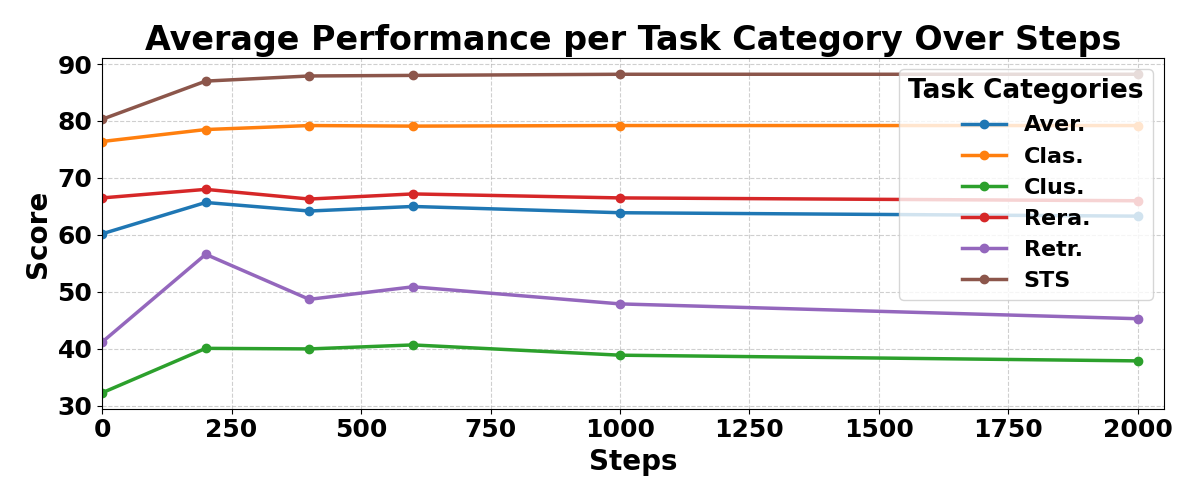}
    \caption{Performance across different training steps.}
    \label{fig:half_epoch_comparison}
\end{figure}

\subsection{Further Analysis}
\paragraph{Contrastive Learning Alleviates Dimensional Collapse} 
As shown in Figure~\ref{fig:pca_dim_second}, the model trained with \ourmethodkl{} + UCL + SCL exhibits a higher effective dimension than the model trained with \ourmethodkl{} + UCL. Furthermore, \ourmethodkl{} + UCL achieves a higher effective dimension than \ourmethodkl{} alone. The reduction of dimensional collapse also correlates with the enhancement of \ourmethod{} over different training stages as shown in the previous section. %For future work, we plan to further investigate how alignment capacity \cite{DBLP:conf/emnlp/GaoYC21} is improved at different training stages of \ourmethod{}.

% As depicted in Figure~\ref{fig:pca_dim_second}, the effective dimension of the model trained with LLM2Comp(KL) + UCL + SCL is higher than that of the model trained with LLM2Comp(KL) + UCL. Furthermore, the effective dimension of LLM2Comp(KL) + UCL is greater than that of LLM2Comp(KL) alone. This indicates that both UCL and SCL tasks enrich the model's representation from LLM2Comp(KL). Although SimCSE~\cite{DBLP:conf/emnlp/GaoYC21} suggests that uniformity loss decreases during unsupervised contrastive learning, it does not demonstrate how the effective dimension changes. While many assume that uniformity loss can reflect the effective dimension, it does not necessarily correlate with an increase in the effective dimension. Moreover, as pointed out in~\cite{DBLP:conf/iclr/Fang0SW24}, uniformity loss, as used in SimCSE, is not an adequate measure of dimensional collapse. In this study, we quantify the effective dimension to show the increase in effective dimension between both the second and third stages.
% \ncamtu{compare effective dimensions (with SVD) before CL, after UCL, after SL}

\paragraph{Convergence Analysis} Models trained with the compression objective exhibit higher data efficiency, achieving peak performance with only 0.36M training samples, compared to the 1.16M samples required by \llmtovec{}, as shown in Table~\ref{tab:two_stage_results}. For a more detailed analysis, Figure~\ref{fig:half_epoch_comparison} shows how the performance of \ourmethod{} evolves over training steps in the SCL stage. We observe that our model converges rapidly, reaching optimal performance within 200 steps and remaining stable for most tasks. However, for retrieval tasks, performance begins to decline when contrastive learning continues beyond this point. \torevise{We hypothesize \ourmethodkl{} can achieve good alignment loss ~\cite{DBLP:conf/icml/0001I20, DBLP:conf/iclr/JingVLT22}, and the subsequent contrastive learning stage rapidly balances effective dimensionality and alignment, leading to faster convergence.} Beyond this point, additional CL adopted in this paper, which uses InfoNCE with fixed negative sampling, becomes less effective.
This phenomenon represents an interesting direction for future research.

% However, further training led to a decrease in performance for tasks other than STS. We propose several hypotheses to explain this unexpected behavior:

% Representation learning optimization may cause the model to forget general knowledge that could have been learned during the pre-training phase, reducing its ability to generalize.

% The model may overfit to the distribution of the original dataset.

% The model may encounter a uniformity-tolerance dilemma, which refers to excessive pursuit to the uniformity makes the contrastive loss not tolerant to semantically similar samples, which may break the underlying semantic structure and be harmful to the formation of features useful for downstream tasks\cite{DBLP:conf/cvpr/WangL21a}.
% If one of the first two hypotheses holds, other models would likely exhibit similar behavior. However, we have not observed such evidence, suggesting that the model is more likely experiencing the uniformity-tolerance dilemma.

\section{Conclusion}

Our study demonstrates the potential of context compression as a pretext task for the unsupervised adaptation of large language models (LLMs). We identify CTKD as the optimal training objective and determine the appropriate number of memory tokens needed for downstream representations. A deeper analysis shows that CTKD effectively mitigates dimensional collapse, resulting in stronger text representations than other pretext tasks. Building on this, additional contrastive learning yields a robust embedding model, LLM2Comp, which outperforms contemporary baselines (LLM2Vec and Llama2Vec) trained with similar recipes but requires much less training data. Furthermore, we provide insights into the effective dimensionality, task-aware performance, and sample efficiency, highlighting promising directions for future research.

\torevise{\section*{Acknowledgements}
This work was supported by the National Natural Science Foundation of China (Grant No. W2532049).}

\bibliography{aaai2026}

\clearpage
\appendix
\section*{Appendix}
\renewcommand{\thesection}{\arabic{section}} 
\section{Limitations and Future Work}
This work demonstrates the potential of context compression as a pretext task for the unsupervised adaptation of large language models. We identify CTKD as an effective training objective and determine the appropriate number of memory tokens needed for downstream representation. Our analysis indicates that CTKD mitigates dimensional collapse, leading to stronger text representations compared to other pretext tasks. Furthermore, incorporating additional contrastive learning yields a robust embedding model, LLM2Comp, which outperforms contemporary baselines (LLM2Vec and Llama2Vec) trained under similar conditions, while requiring substantially less training data.  
Despite these promising results, our conclusions are primarily based on empirical evidence. A deeper theoretical analysis is needed to formally establish the connection between the InfoNCE loss and dimensional collapse in both unsupervised and supervised contrastive learning.

\begin{table*}
\scriptsize
\centering
\begin{tabular}{p{0.23\linewidth}p{0.72\linewidth}}
\toprule
\textbf{Dataset} & \textbf{Instruction(s)} \\
\midrule
DuReader & Given a Chinese search query, retrieve web passages that answer the question \\
ELI5 & Provided a user question, retrieve the highest voted answers on Reddit ELI5 forum \\
FEVER & Given a claim, retrieve documents that support or refute the claim \\
HotpotQA & Given a multi-hop question, retrieve documents that can help answer the question \\
MIRACL & Given a question, retrieve Wikipedia passages that answer the question \\
MrTyDi & Given a question, retrieve Wikipedia passages that answer the question \\
MSMARCO Document & Given a web search query, retrieve relevant documents that answer the query \\
MSMARCO Passage & Given a web search query, retrieve relevant passages that answer the query \\
NLI & Given a premise, retrieve a hypothesis that is entailed by the premise \\
    & Retrieve semantically similar text \\
NQ & Given a question, retrieve Wikipedia passages that answer the question \\
QuoraDuplicates & Given a question, retrieve questions that are semantically equivalent to the given question \\
 & Find questions that have the same meaning as the input question \\
SQuAD & Retrieve Wikipedia passages that answer the question \\
T2Ranking & Given a Chinese search query, retrieve web passages that answer the question \\
TriviaQA & Retrieve Wikipedia passages that answer the question \\
\bottomrule
\end{tabular}
\caption{Instructions for finetuning E5 datasets.}
\label{tab:finetuning_instructions}
\end{table*}
\section{Implementation Details}
\subsection{Training Details}
\label{implement_details}
\subsubsection{Compression Pretext Training Details}
We expanded the vocabulary by adding 8 special tokens, which increased the vocabulary size from 32,000 to 32,008. Accordingly, the embedding layer was reshaped to match the new dimensions. The number of memory tokens was selected based on the experimental analysis presented in Section~\ref{token_length_impact}. Following the approach of LLM2Vec, we employed LoRA (Low-Rank Adaptation) for efficient parameter fine-tuning. Specifically, the LoRA rank was set to 16, and the alpha parameter was set to 32 based on empirical evidence and consistent with \llmtovec{}. The modules modified by LoRA include the query, value, and output projections within the attention layer, as well as the up, down, and gate projections within the feedforward network layers. Since the newly added special tokens were specifically introduced to compress and encode semantic information, we kept the embedding layer fully trainable to ensure that the model could effectively adapt to the representation learning space.

The model was trained for 8,000 steps with a batch size of 4, resulting in a total of 32,000 samples from English Wikipedia, consistent with \llmtovec{}. We selected Wikipedia because it is presumably included in the pre-training corpus of the model used in our experiments. Thus, this adaptation step is not expected to provide new factual knowledge but rather to refine the model’s ability to compress sentences and construct sequence representations, making the comparison with \llmtovec{} appropriate. Specifically, we used the Wikitext-103 dataset \cite{DBLP:conf/iclr/MerityX0S17} for training. During training, we utilized bfloat16 precision to optimize memory usage. The base model was the “meta-llama/Llama-2-7b-chat-hf”\footnote{https://huggingface.co/meta-llama/Llama-2-7b-chat-hf}. We set the learning rate to 1e-4 and the weight decay to 1e-5, parameters chosen for stable training loss. We also applied DeepSpeed ZeRO-0 optimization training, along with a warm-up decay learning rate schedule, where the minimum learning rate during warm-up was set to 1e-5. To examine the impact of random seed selection on model performance, we report results for two representative cases: a seed (2026) that consistently led to stronger performance and a seed (42) that resulted in weaker performance. For all experiments, random seeds were fixed across PyTorch, NumPy, and Python’s random library to ensure reproducibility within each setting.
\subsubsection{Unsupervised Contrastive Learning Training Details}
Following the SimCSE\cite{DBLP:conf/emnlp/GaoYC21}, we use dropout to get the unsupervised positive samples, and in-batch sentences are regarded as negative samples, and the InfoNCE loss was then applied for unsupervised contrastive learning. The dropout rate was set to 0.2, and the batch size was 128 with gradient checkpointing. The criterion used for selecting dropout rate is stable training loss and batch size is selected to be consistent with \llmtovec{}. We employed LoRA (Low-Rank Adaptation) for efficient parameter fine-tuning. Specifically, the LoRA rank was set to 16, and the alpha parameter was set to 32 based on empirical evidence and consistent with \llmtovec{}. The modules modified by LoRA include the query, value, and output projections within the attention layer, as well as the up, down, and gate projections within the feedforward network layers.

The model was trained for 1,000 steps with a batch size of 128, resulting in a total of 128,000 samples from English Wikipedia on a single H800, also consistent with \llmtovec{}. We selected Wikipedia because it is presumably included in the pre-training corpus of the model used in our experiments. Thus, this adaptation step is not expected to provide new factual knowledge but rather to refine the model’s ability to compress sentences and construct sequence representations, making the comparison with \llmtovec{} appropriate. Specifically, we used a subset of Wikipedia sentences released by \cite{DBLP:conf/emnlp/GaoYC21} for training. During training, we utilized bfloat16 precision to optimize memory usage.  The learning rate was set to 3e-5, with weight decay of 1e-3. We also applied DeepSpeed ZeRO-0 optimization, along with a warm-up decay learning rate schedule, where the minimum learning rate during warm-up was set to 1e-5. These parameters are chosen for the stable training loss.
\begin{table}[t]
    \centering
    \small
    \resizebox{0.95\columnwidth}{!}{
    \begin{tabular}{c|c|c}
    \toprule
    \textbf{Category} & \textbf{Dataset}  & \textbf{\#Samples }\\
    \midrule
    \multirow{3}{*}{Clustering (3)} & BiorxivCS2S & 75000\\
    & MedrxivS2S  & 37500\\
    & \makecell{TwentyNewsgroups}  & 59545\\ \midrule
    \multirow{3}{*}{Retrieval (3)} & SciFact & 5483  \\
    & NFCorpus & 3956\\
    & ArguAna  & 10080\\ \midrule
    \multirow{3}{*}{STS (3)} & STS17 & 5692\\
    & SICK-R  & 19854\\
    & STSBenchmark & 2758\\ \midrule
    \multirow{3}{*}{\makecell{(Pair)\\ Classification (3)}} & Banking77 & 3696\\
    & EmotionClassification & 2096\\
    & SprintDuplicateQuestions & 8931\\ \midrule
    \multirow{2}{*}{Reranking (2)} & \makecell{StackOverflow\\DupQuestions} & 82798\\
    & SciDocsRR & 89131\\ \midrule
    Overall & 14 datasets & 406520\\
    \bottomrule
    \end{tabular}}
    \caption{Statistics of evaluation datasets}
    \label{tab:mteb-subset}
\end{table}
\begin{table*}
\centering   
\scriptsize
\begin{tabular}{p{0.24\linewidth}p{0.7\linewidth}}
\toprule
\textbf{Task Name} & \textbf{Instruction} \\
\midrule
ArguAna & Given a claim, find documents that refute the claim  \\
Banking77Classification & Given an online banking query, find the corresponding intents  \\
BiorxivClusteringS2S & Identify the main category of Biorxiv papers based on the titles  \\
EmotionClassification &  Classify the emotion expressed in the given Twitter message into one of the six emotions: anger, fear, joy, love, sadness, and surprise  \\
MedrxivClusteringS2S & Identify the main category of Medrxiv papers based on the titles  \\
NFCorpus & Given a question, retrieve relevant documents that best answer the question  \\
SciDocsRR & Given a title of a scientific paper, retrieve the titles of other relevant papers  \\
SciFact & Given a scientific claim, retrieve documents that support or refute the claim  \\
StackOverflowDupQuestions & Retrieve duplicate questions from StackOverflow forum  \\
SICK-R & Retrieve semantically similar text.  \\
SprintDuplicateQuestions & Retrieve duplicate questions from Sprint forum  \\
STS17 & Retrieve semantically similar text.  \\
STSBenchmark & Retrieve semantically similar text.  \\
TwentyNewsgroupsClustering & Identify the topic or theme of the given news articles  \\
\bottomrule
\end{tabular}
\caption{Instructions used for our evaluation datasets.} \label{tab:mteb_subset_instructions}
 %\va{shift to the end}}
\end{table*}
\subsubsection{Supervised Contrastive Learning Training Details}
Following \llmtovec{} \cite{DBLP:journals/corr/abs-2404-05961}, we used the E5 dataset for training. This dataset consists of ELI5 (sample ratio 0.1) \citep{DBLP:conf/acl/FanJPGWA19}, HotpotQA \citep{yang-etal-2018-hotpotqa}, FEVER \citep{thorne-etal-2018-fever}, MIRACL \citep{zhang-et-all-2023-MIRACL}, MS-MARCO passage ranking (sample ratio 0.5) and document ranking (sample ratio 0.2) \citep{DBLP:conf/nips/NguyenRSGTMD16}, NQ \citep{karpukhin-etal-2020-DPR}, NLI \citep{DBLP:conf/emnlp/GaoYC21}, SQuAD \citep{rajpurkar-etal-2016-squad},
TriviaQA \citep{joshi-etal-2017-triviaqa},
Quora Duplicate Questions (sample ratio 0.1),
Mr- TyDi \citep{zhang-etal-2021-mr}, DuReader \citep{he-etal-2018-dureader}, and T2Ranking (sample ratio 0.5) \citep{t2ranking}. This is a public dataset widely used by \llmtovec{} \cite{DBLP:journals/corr/abs-2404-05961}, mE5 \cite{DBLP:journals/corr/abs-2402-05672}, $E5_{mistral}$ \cite{DBLP:conf/acl/WangYHYMW24}, GritLM \cite{DBLP:journals/corr/abs-2402-09906} and so on. The fine-tuning instructions for each dataset are the same as those used in \llmtovec{}, and are provided in Table~\ref{tab:finetuning_instructions}. 

The model was trained for 200 steps with a batch size of 128 on 8 NVIDIA H800 GPUs, yielding an effective batch size of 1,024 and processing approximately 128,000 training instances from the E5 dataset. As the E5 dataset is a widely adopted benchmark for supervised contrastive learning, using it enables a fair comparison with many existing models. Specifically, we used a subset of the E5 dataset for training. During training, we employed bfloat16 precision to reduce memory usage. The learning rate was set to 1e-4 with a weight decay of 3e-4. We also applied DeepSpeed ZeRO-0 optimization, together with a warm-up decay learning rate schedule, where the minimum learning rate during warm-up was set to 1e-5. These hyperparameters were selected to ensure stable training loss.

Here, we provide brief introductions to the baseline models compared in our main experiments:
\label{compared_methods}
We compare our approach against several strong baselines: 

\noindent\textbf{Instructor}~\cite{DBLP:conf/acl/SuSKWHOYSZ023}: 
An embedding model that introduces instruction tuning, extending GTR~\cite{DBLP:conf/emnlp/Ni0LDAMZLHCY22}, and leveraging a curated dataset spanning a wide range of tasks.  

\noindent\textbf{ULLME}~\cite{DBLP:journals/corr/abs-2408-03402}: 
An approach based on Generation-Representation Learning (GRL), which jointly optimizes contrastive learning and generation objectives.  

\noindent\textbf{BGE-ICL}~\cite{DBLP:conf/iclr/LiQXCLLSL25}: 
An embedding model that leverages the in-context learning (ICL) capabilities of large language models to enhance embedding quality.  

\noindent\textbf{RepLlama}~\cite{DBLP:conf/sigir/MaWYWL24}: 
A fine-tuned LLaMA-2-7B model, optimized for multi-stage text retrieval tasks.

In the main body of the paper, the number of training samples for \llmtovec{} was mistakenly reported as 1.16 million due to a typographical error. The correct total is 1.66 million, comprising 1.5 million samples from the supervised contrastive learning stage (8 GPUs, 64 micro-batches, 1000 steps, 3 epochs) and 0.16 million samples from the pretext and unsupervised contrastive learning stages. This correction does not affect the conclusions of this paper. We apologize for the oversight and any confusion it may have caused.
%The instruction used for each dataset can be found in \Cref{tab:appendix:finetuning_instructions}.
\subsection{Evaluation Details}
\label{evaluation_detail}
Our evaluation data are presented in Table~\ref{tab:mteb-subset}, which lists the subset of the full MTEB benchmark used in our experiments. To balance resource expenditure with evaluation coverage and to accelerate the evaluation process, we selected 14 datasets, consistent with those included in \llmtovec{}. To ensure that our ablation studies and analyses are not biased toward a particular category or task, this subset was constructed to include tasks from each category in proportions that approximately match those in the full MTEB benchmark. During evaluation, all models were provided with the same instructions as in \cite{DBLP:conf/acl/WangYHYMW24} and \cite{DBLP:journals/corr/abs-2404-05961}. The evaluation metrics followed the MTEB standard \cite{DBLP:conf/eacl/MuennighoffTMR23}: Accuracy for classification tasks, V-measure for clustering, NDCG@10 for retrieval, MAP for reranking, and Spearman correlation for STS. We followed the instructions of \llmtovec{}, shown in Table~\ref{tab:mteb_subset_instructions}.
\subsection{The Computing Infrastructure}
All training and evaluation experiments were conducted on NVIDIA H800 GPUs running Ubuntu 22.04 x86\_64 with a system memory capacity of 666 GB. The compression pretext task and unsupervised contrastive learning were performed on a single NVIDIA H800 GPU, whereas supervised contrastive learning was conducted across 8 NVIDIA H800 GPUs.

\section{Further Analysis}

\begin{figure}
    \centering
    \includegraphics[width=0.95\linewidth]{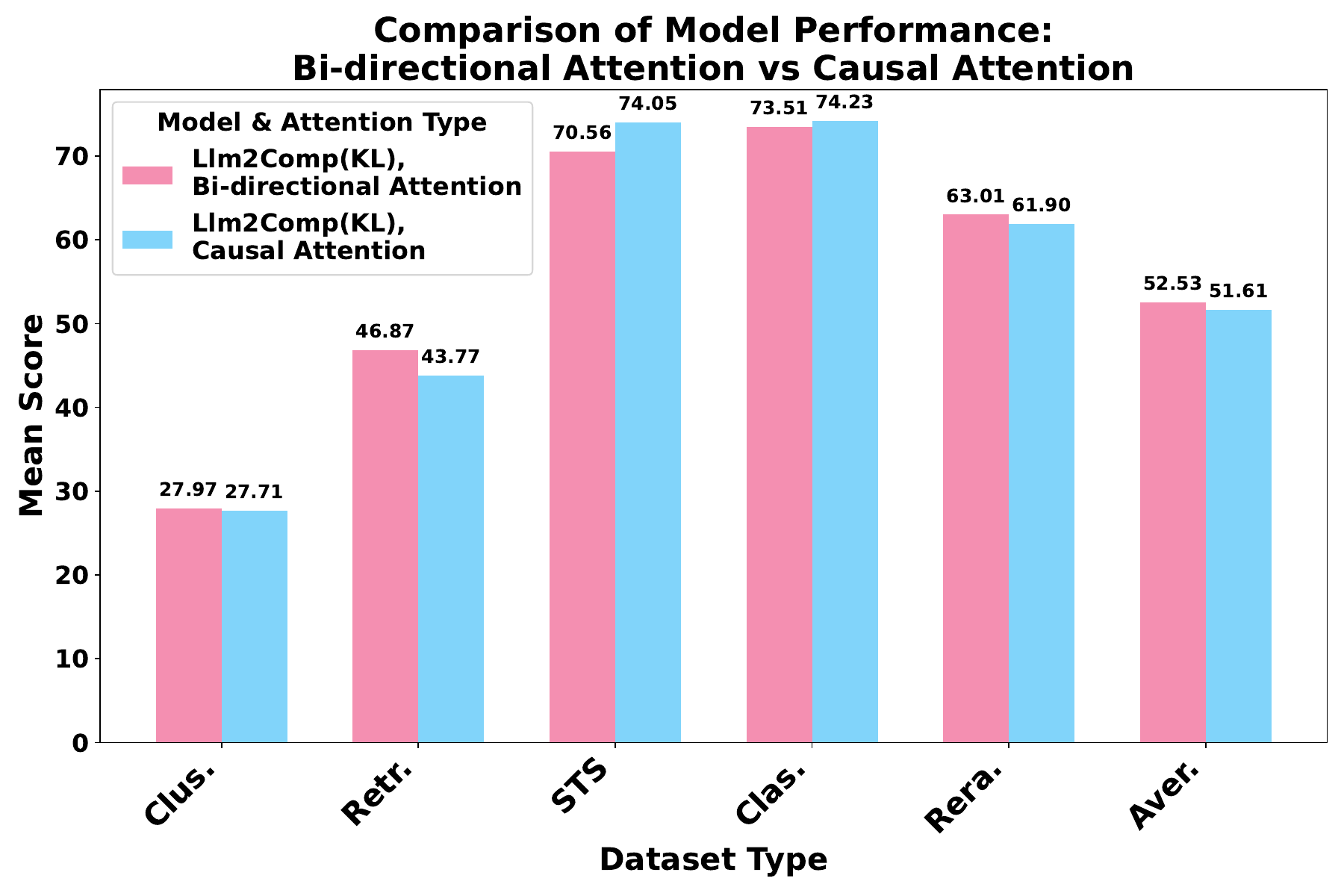}
    \caption{\torevise{Comparison of model performance across different data types (Clustering, Retrieval, STS, Classification, and Reranking) with two attention mechanisms: Bi-directional and Causal attention.}}
    \label{fig:bi_vs_causal}
\end{figure}
\begin{figure}[htbp]
\centering
\resizebox{0.9\linewidth}{!}{
\includegraphics{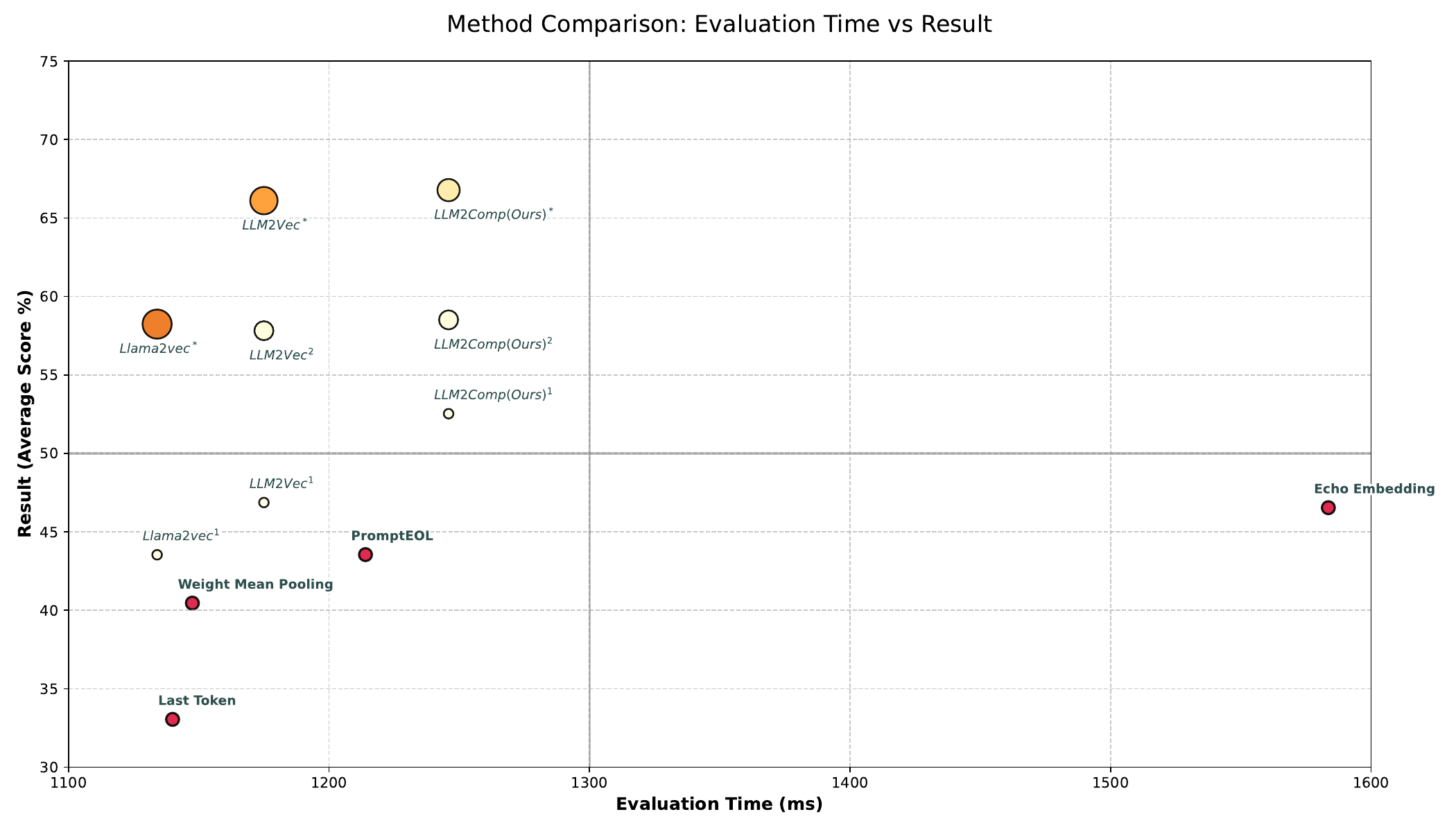}
}
\caption{\small \torevise{Comparison of inference time and performance.}}
\label{fig:comparison_time}
\end{figure}

\label{attention_architecture}
\paragraph{Bidirectional vs Causal Architecture}
We examine the impact of adapting causal attention to bidirectional attention on overall performance. To this end, we train \ourmethodkl{} using causal LLMs and, for comparison, using the same training recipe with a bidirectional architecture described in Section \ref{sec:rep-via-comp}. As shown in Figure~\ref{fig:bi_vs_causal}, the bidirectional variant outperforms its causal counterpart on average, consistent with prior findings \cite{DBLP:journals/corr/abs-2402-09906}. However, the advantage of bidirectional attention is pronounced in retrieval tasks, whereas the causal variant is more beneficial for  STS and classification tasks. This observation highlights the importance of selecting the appropriate architecture based on the target downstream application.
\torevise{\paragraph{Inference Time} We sample 10\% of the data from each evaluation dataset to estimate the inference time for all models. This approach allows us to gauge the computational efficiency of each model across different datasets. In Figure~\ref{fig:comparison_time}, red dots represent training-free methods, while larger marker sizes and darker colors indicate models that use progressively more training samples. The figure provides a visual comparison of inference time against model performance, highlighting the trade-offs between these two aspects for each method.}

\torevise{In this figure, red markers represent training-free methods, while the color gradient (from light to dark) and marker size (from small to large) indicate the increasing amount of training data used by the models. Notably, except for Echo Embedding, the time overhead for all other methods does not differ substantially. This suggests that additional training data and the use of a LoRA architecture do not substantially affect the inference time for most models. In contrast, our method achieves the best performance while requiring fewer training samples than both \llmtovec{} and \lamatovec{}. This demonstrates that our approach strikes a better balance between performance and data efficiency.}

\end{document}